\documentclass{ieeeaccess}

\usepackage{cite}
\usepackage{amsmath,amssymb,amsfonts}
\usepackage{graphicx}
\usepackage{textcomp}
\usepackage{booktabs}
\usepackage{url}

\def\BibTeX{{\rm B\kern-.05em{\sc i\kern-.025em b}\kern-.08em
    T\kern-.1667em\lower.7ex\hbox{E}\kern-.125emX}}

\providecommand{\doi}[1]{}

\begin{document}

\doi{10.1109/ACCESS.XXXX.XXXXXXX}

\title{BRUCE: Benchmarking Robustness Under Corruption Escalation for Scientific Vision-Language Reasoning}

\author{\uppercase{Saim Rehman (Student Member, IEEE)}\authorrefmark{1}, AND
\uppercase{Muhammad Shafique (Senior Member, IEEE)}\authorrefmark{1}}

\address[1]{eBRAIN Lab, New York University Abu Dhabi (NYUAD), Abu Dhabi, United Arab Emirates
(e-mail: sr7849@nyu.edu; muhammad.shafique@nyu.edu)}

\markboth
{Rehman and Shafique: BRUCE: Benchmarking Robustness Under Corruption Escalation}
{Rehman and Shafique: BRUCE: Benchmarking Robustness Under Corruption Escalation}

\corresp{Corresponding author: Saim Rehman (e-mail: sr7849@nyu.edu).}

\tfootnote{
This work was supported in part by the NYUAD Center for CyberSecurity (CCS), funded by Tamkeen under the NYUAD Research Institute Award G1104.}

\begin{abstract}
Visual-language models (VLMs) frequently struggle with robustness issues in real-world situations due to low-/varying-quality input images. 
In this paper, we aim at analyzing VLMs' robustness by applying perturbations and distortions to the input images, such as blur or low contrast. 
Towards this, we propose \textit{BRUCE}, \textbf{B}enchmarking \textbf{R}obustness \textbf{U}nder \textbf{C}orruption \textbf{E}scalation, a multimodal reasoning fragility framework for scientific vision-language reasoning. 

State-of-the-art evaluation frameworks/studies primarily focus on clean-task accuracy and rarely analyze how reasoning stability degrades across robustness dimensions. 
Besides varying over a wide-range of input perturbations, BRUCE employs two novel metrics -- Robustness Corruption Index (RCI) and Traversal-RCI (T-RCI) -- to quantify how rapidly multimodal reasoning performance deteriorates in VLMs as visual corruption severity increases under progressive perturbation scaling.
We evaluate BRUCE across chemistry and mathematical reasoning tasks for multiple datasets, while analyzing corruption-induced prediction failures in terms of four high-level reasoning domains: OCR-dependent reasoning, spatial reasoning, symbolic reasoning, and semantic failures, with each containing fine-grained corruption-specific failure subtypes, thereby enabling an interpretable failure analysis. 

\end{abstract}


\maketitle

\section{Introduction}

Vision-Language Models (VLMs) have emerged as a major research direction to study how multimodal systems can integrate visual and linguistic information for tasks including visual question answering, scene understanding, scientific reasoning, document analysis, robotics, and embodied decision making~\cite{liu2023llava,bai2023qwenvl,chen2024internvl,dai2023instructblip,wang2023cogvlm}. Increasing capable visual understanding and reasoning abilities in modern VLMs have enabled them to support applications that require multimodal interpretation and reasoning across scientific and real-world contexts. 
However, despite the strong clean-task performance of VLMs on a wide range of multimodal reasoning benchmarks, their robustness and reliability remain poorly understood (in real-world settings) under degraded inputs, visual perturbed conditions and domain shifts ~\cite{recht2018cifar, recht2019imagenet}.
When input distortions/corruptions (with varying degree) such as blur, rotation, contrast degradation, and cropping are introduced, visual information is substantially altered/degraded, and in downstream reasoning tasks, disproportionate failures are induced. 
In scientific and mathematical reasoning settings, these perturbations can distort diagrams, symbolic structures, molecular representations, or geometric relationships, potentially leading to incorrect multimodal reasoning behavior despite relatively minor changes in image quality. 

\subsection{State-of-the-Art and Their Major Limitations}

A large body of related work, which focuses primarily on clean-task accuracy and benchmark-level reasoning performance \cite{liu2023mmbench}\cite{yue2023mmmu}\cite{chen2024mmstar}\cite{ying2024mmtbench}, does not provide a comprehensive analysis of reasoning stability degradation across different robustness dimensions. Previous work shows that robustness is inherently multi-faceted and, moreover, cannot be adequately characterized by a single benchmark or distribution shift~\cite{hendrycks2019benchmarking}. Several multimodal robustness frameworks/studies \cite{hendrycks2019benchmarking, liu2024rbench,
agarwal2025vqa,
usama2025corruptions,
sui2025benchc, arXiv:2412.19794}
evaluate robustness degradation using  \textit{aggregated corruption-response measures} such as Absolute Robustness and Relative Robustness \cite{liu2024rbench}, or \textit{severity-aggregated metrics} such as Average Error, Error Rate, Range of Error, and Visual Robustness Error \cite{agarwal2025vqa}. 
While these evaluations provide useful estimates of performance degradation under selected cases of input corruption, including chemistry-oriented multimodal reasoning benchmarks such as ChemVLM~\cite{li2025chemvlm}, ChemTable~\cite{zhou2025chemtable}, and MolPuzzle~\cite{guo2024molpuzzle}, they largely focus on aggregate predictive robustness and \textit{do not analyze how perturbations propagate into incorrect scientific reasoning predictions}.

Prior robustness benchmarks such as MVTamperBench \cite{arXiv:2412.19794}, evaluate binary tampering detection performance using predictive metrics such as F1 scores, which may not provide comprehensive robustness insights and reasoning, and interpretable failure analysis. 
Recent studies of corruption-focused robustness mainly analyze perception-level degradation for low-level vision and detection tasks under calibrated corruption settings \cite{becker2026causally}, and lack a wide-range of input degradations/perturbations. 
Furthermore, these approaches often fail to capture how visual distortions propagate into semantic reasoning failures in image-grounded tasks. Unlike Bench-C~\cite{sui2025benchc}, which extends robustness evaluation beyond accuracy through prediction-structure and calibration-aware metrics, it does not analyze how corruption induced perturbations propagate into distinct multimodal reasoning failure modes or directional reasoning fragility. 

\subsection{Our Novel Contributions}

To address the above challenges, we propose the following novel contributions:

\vspace{5pt}
\begin{enumerate}
\item We introduce \textit{BRUCE, a multimodal reasoning fragility framework for scientific vision-language reasoning}, that performs comprehensive perturbation-scaling evaluations for analyzing corruption-induced reasoning failures in VLMs across chemistry and mathematical reasoning benchmarks.
\vspace{5pt}
\item We empower BRUCE through two novel metrics -- \textit{Robustness Corruption Index (RCI) and Traversal-RCI (T-RCI)} -- that quantify how rapidly multimodal reasoning performance
deteriorates under visual corruption and progressive spatial information loss. RCI measures robustness degradation under increasing corruption severity by jointly modeling prediction accuracy loss, calibration drift, entropy instability, and failure propagation, whereas T-RCI extends this formulation to traversal perturbations by additionally capturing directional sensitivity, spatial evidence removal, and progression-induced reasoning collapse. Together, these metrics enable systematic evaluation of both corruption robustness and spatial reasoning robustness in VLMs.

\vspace{5pt}
\item We introduce \textit{a controlled perturbation scaling protocol} that spans progressive blur, rotation, contrast degradation, and different levels of spatial cropping severity, enabling systematic analysis of robustness degradation trajectories beyond conventional single-step corruption evaluation. 
We further extend the framework towards \textit{progressive spatial evidence removal} via top-down, bottom-up, left-to-right, and right-to-left traversals, allowing robustness to be assessed under directional information loss and spatial reasoning degradation.
\vspace{5pt}
\item Through extensive robustness analyses across \textit{ScienceQA Chemistry} \cite{lu2022learn}, \textit{MMJEE Chemistry} \cite{mukherjee2025mmjee}, and \textit{MathVista MCQ} \cite{lu2023mathvista}, we show that strong clean-task accuracy does not reliably predict robustness under corruption escalation and that different perturbation families induce distinct failure modes/classes (as discussed below).

\vspace{5pt}

\item To support fine-grained multimodal reasoning analysis, we introduce \textit{a hierarchical failure taxonomy comprising four high-level failure classes}: OCR-dependent reasoning, spatial reasoning, symbolic reasoning, and semantic failures. Each failure class is further decomposed into fine-grained corruption-specific subtypes. The resulting failure taxonomy exposes reasoning breakdown mechanisms that remain hidden under aggregate robustness metrics, enabling systematic diagnosis of model vulnerabilities and corruption-specific failure patterns.

\end{enumerate}

\section{BRUCE: VLM Robustness Benchmarking Framework}

\begin{figure*}[!t]
\centering
\includegraphics[width=\textwidth]{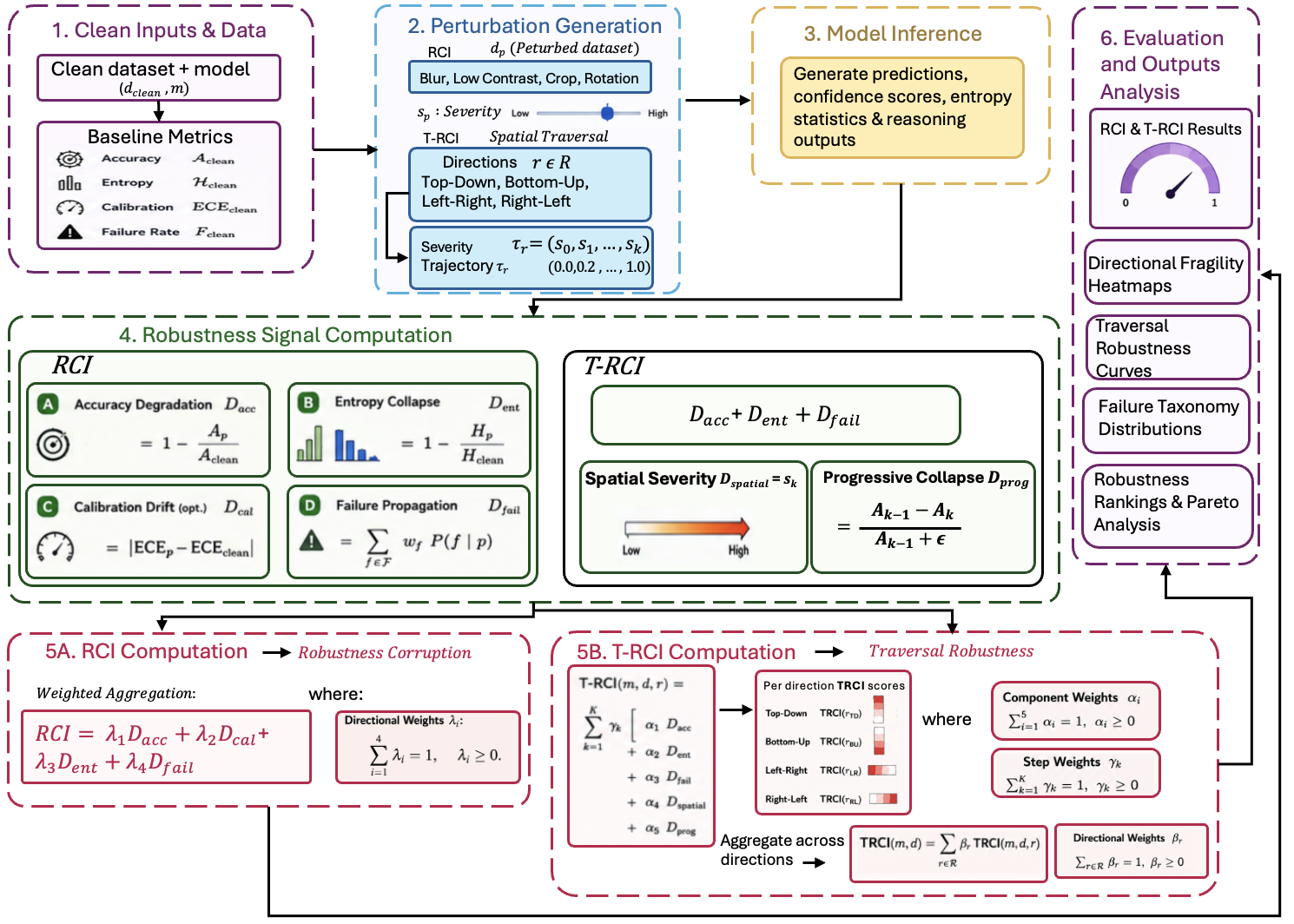}
\caption{
A Comprehensive Overview of BRUCE Framework illustrating Different Steps
}
\label{fig:BRUCEoverview}
\end{figure*}

Figure~\ref{fig:BRUCEoverview} provides a step-by-step overview of our BRUCE framework, illustrating different key features of the framework. BRUCE performs the following key steps (discussed in detail in the following sub-sections).
\vspace{5pt}
\begin{enumerate}
    \item \textbf{Clean Inputs and Data:} First, clean samples and ground-truth labels are collected to establish the baseline reference for robustness evaluation under unperturbed conditions.
    \vspace{5pt}
    \item \textbf{Perturbation Generation:} The input corruption conditions and spatial traversal trajectories are generated by progressively applying perturbations across multiple severity levels and traversal directions.
    \vspace{5pt}
    \item \textbf{Model Inference:} 
    
    For each given VLM model, inference is performed on perturbed and traversal-modified samples (from different datasets) while recording predictions, confidences, reasoning traces, and correctness signals.
    \vspace{5pt}
    \item \textbf{Robustness Signal Computation:} To analyze degradation in  model robustness, different metrics like predictive degradation, entropy collapse, calibration drift, failure propagation, and traversal progression signals are computed for the VLM output under perturbed samples.

    \vspace{5pt}
    \item \textbf{RCI and T-RCI Computation:} To achieve consolidated robustness quantification, combined effects of different types of degradation are integrated in RCI and T-RCI metrics. 
    RCI aggregates robustness degradation under individual perturbation conditions, while T-RCI aggregates progressive robustness collapse across traversal severity trajectories and directions.
    \vspace{5pt}
    \item \textbf{Evaluations and Output Analysis:} To derive valuable observations from our benchmark studies, the aggregated robustness metrics (RCI, T-RCI) are meticulously analyzed using heatmaps, degradation curves, failure taxonomies, collapse dynamics, and comparative robustness rankings across different models and datasets.
\end{enumerate}
\vspace{-10pt}

\subsection{Clean Inputs and Data}

Given an image-question pair \((x,q)\) from a benchmark dataset \(d\),  BRUCE first evaluates the vision-language model (VLM) \(m\) under clean conditions to establish baseline multimodal reasoning performance. The clean-condition prediction is defined as 
$\hat{y}_{m,d,\mathrm{clean}} = f_m(x,q)$, 
where \(f_m(\cdot)\) denotes the multimodal inference behavior of model \(m\).

\subsection{Perturbation Generation}

\begin{figure*}[!t]
\centering
\includegraphics[width=\textwidth]{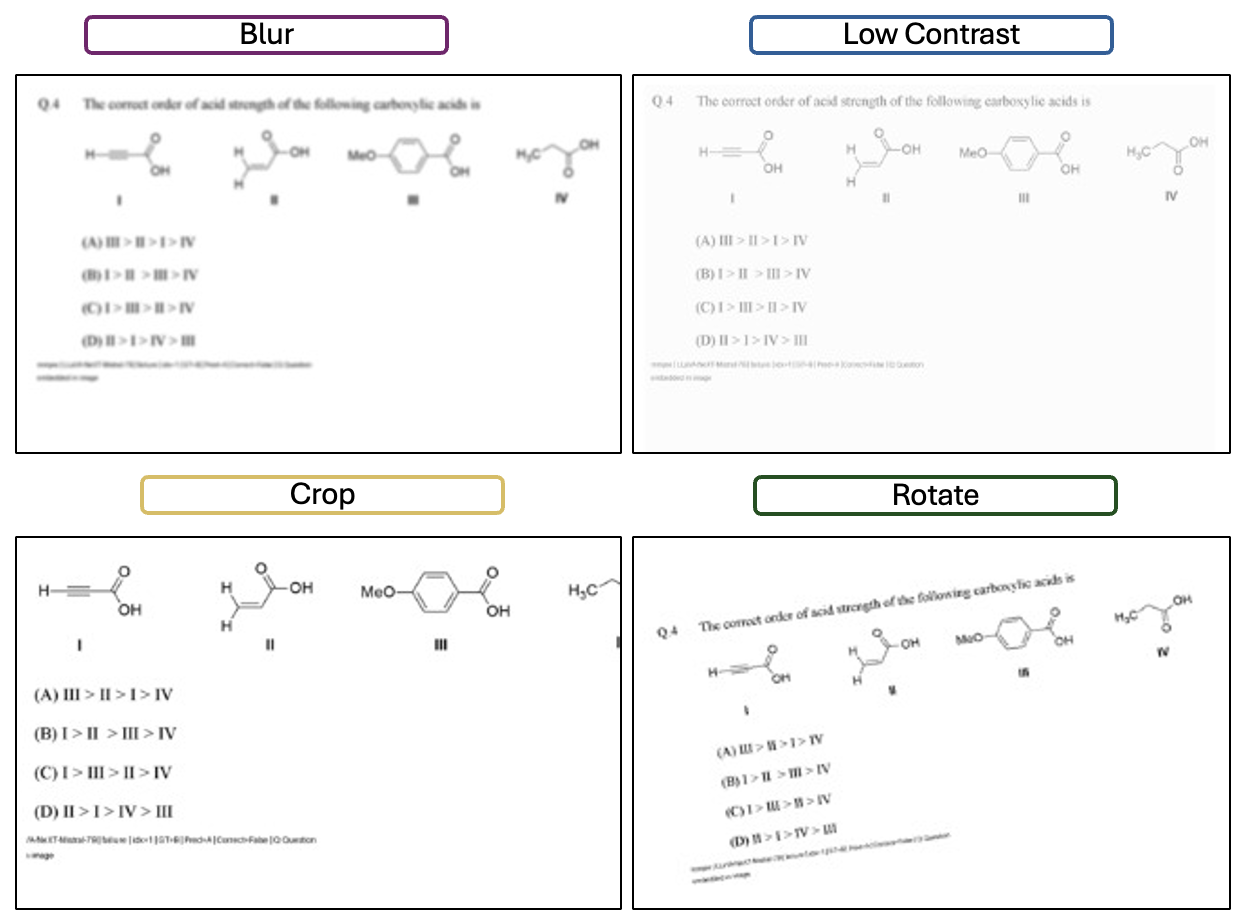}
\caption{
Examples of Single Perturbation Types: (a) Blurry Image; (b) Low-Contrast Image; (c) Cropped Image; and (d) Rotated Image.
}
\label{fig:example}
\end{figure*}

BRUCE evaluates robustness under both single-condition perturbations and traversal-based corruption escalation. Let $x_p=\mathcal{T}_p(x,s)$
denote a perturbed image generated by applying perturbation type \(p\) at severity level \(s\) using corruption operator \(\mathcal{T}_p\). Perturbation categories considered in this work include \textit{Gaussian blur, low contrast, center crop, and image rotation}. Figure~\ref{fig:example} illustrates examples of single perturbation types.

Regarding T-RCI, let \(r\in\mathcal{R}\) denote a spatial traversal
direction, where
\[
\begin{aligned}
\mathcal{R}=\{&
\text{topdown},\ \text{bottomup},\\
&
\text{leftright},\ \text{rightleft}
\}.
\end{aligned}
\]
Let
\(\tau_r=(s_0,s_1,\dots,s_k)\)
represent the ordered traversal severity trajectory associated with
direction \(r\), where \(s_0=0\) corresponds to the clean condition and
\(s_k\) denotes progressively increasing traversal severity.

\subsection{Model Inference}

For each traversal state \(s_k\), the model produces a traversal-conditioned prediction
\vspace{5pt}
\[
\hat{y}_{m,d,r,s_k}
=
f_m(\mathcal{T}_r(x,s_k),q),
\]
where \(\mathcal{T}_r\) denotes the traversal corruption operator associated with direction \(r\).

Unlike conventional robustness evaluations that primarily measure predictive degradation under isolated corruption settings, BRUCE jointly evaluates corruption and traversal robustness through predictive degradation, calibration drift, entropy collapse, perturbation-induced failure propagation, and progressive robustness collapse dynamics. 
\textit{This formulation enables the analysis of degradation trajectories, directional fragility, and multimodal reasoning collapse behavior under progressively increasing corruption severity rather than relying solely on binary clean-versus-corrupted evaluation settings.}

For each perturbation and traversal condition, BRUCE computes predictive accuracy, prediction entropy, failure transition statistics, directional robustness trends, and corruption-induced reasoning degradation dynamics. Calibration-sensitive robustness signals, including Expected Calibration Error (\(ECE\)) and calibration drift, are additionally evaluated when confidence outputs are available for the underlying model.

\subsection{Robustness Signal Computation}

\subsubsection{RCI Robustness Signals}

\noindent For model $m$, dataset $d$, and perturbation condition $p$, \textit{Predictive Degradation}, $\mathcal{D}_{acc}(m,d,p)$, measures the relative reduction in predictive accuracy under corruption, and it is defined as:
\vspace{5pt}
\begin{equation}
\mathcal{D}_{acc}(m,d,p)
=
1-
\frac{
A_{m,d,p}
}{
A_{m,d,\mathrm{clean}}
},
\end{equation}

where \(A_{m,d,p}\) denotes perturbed accuracy and \(A_{m,d,\mathrm{clean}}\) denotes clean-condition accuracy.

\noindent \textit{Calibration Degradation}, $\mathcal{D}_{cal}(m,d,p)$, is defined as Expected Calibration Error (ECE) drift:
\vspace{5pt}
\begin{equation}
\mathcal{D}_{cal}(m,d,p)
=
\left|
\mathrm{ECE}_{m,d,p}
-
\mathrm{ECE}_{m,d,\mathrm{clean}}
\right|,
\end{equation}

when confidence outputs are available for the evaluated model.

When calibration confidence is unavailable, $\mathcal{D}_{cal}$ is not treated as zero degradation; instead, RCI is normalized over the available components only. Thus, a model without calibration is averaged over three terms rather than four, avoiding unnecessary forced penalization or artificial benefit from missing confidence outputs. All models remain directly comparable because missing calibration signals neither penalize nor advantage any model.

\noindent \textit{Entropy Collapse}, $\mathcal{D}_{ent}(m,d,p)$, measures reduction in prediction diversity under corruption:
\vspace{-5pt}
\begin{equation}
\mathcal{D}_{ent}(m,d,p)
=
1-
\frac{
H_{m,d,p}
}{
H_{m,d,\mathrm{clean}}
},
\end{equation}

where \(H_{m,d,p}\) denotes normalized prediction entropy under perturbation.

The entropy term $\mathcal{D}_{ent}$ is not based on internal model probabilities; it is computed from the empirical distribution of the model predictions over the evaluation set under each perturbation condition. Consequently, entropy can be computed consistently for both open-source and API-based models without requiring access to logits or token probabilities.

\noindent \textit{Failure Propagation Severity}, $\mathcal{D}_{fail}(m,d,p)$, is computed as:
\begin{equation}
\mathcal{D}_{fail}(m,d,p)
=
\sum_{f \in \mathcal{F}}
w_f
P
\left(
f
\middle|
\text{clean-correct}
\rightarrow
\text{perturbed-wrong}
\right),
\end{equation}

where \(\mathcal{F}\) denotes the set of failure categories and \(w_f\) denotes the severity weight associated with failure type \(f\).

\subsubsection{T-RCI Traversal Robustness Signals}

\noindent \textit{Traversal-Robustness Corruption Index (T-RCI)} extends robustness analysis to progressive spatial traversal corruption trajectories. 

\noindent \textit{Traversal Predictive Degradation}, $\mathcal{D}_{acc}(m,d,r,s_k)$, is defined as:

\begin{equation}
\mathcal{D}_{acc}(m,d,r,s_k)
=
\max
\left(
0,
\frac{
A_{m,d,r,0}
-
A_{m,d,r,s_k}
}{
A_{m,d,r,0}
}
\right),
\end{equation}

\(A_{m,d,r,s_k}\) denotes traversal accuracy at traversal state \(s_k\). \(A_{m,d,r,0}\) denotes clean accuracy.

\noindent \textit{Traversal Entropy Collapse}, $\mathcal{D}_{ent}(m,d,r,s_k)$, is defined as:

\begin{equation}
\mathcal{D}_{ent}(m,d,r,s_k)
=
\max
\left(
0,
\frac{
H_{m,d,r,0}
-
H_{m,d,r,s_k}
}{
H_{m,d,r,0}
}
\right),
\end{equation}

where \(H_{m,d,r,s_k}\) denotes normalized prediction entropy at traversal state \(s_k\).

\noindent \textit{Traversal-induced Failure Propagation}, $\mathcal{D}_{fail}(m,d,r,s_k)$, is computed as:

\begin{equation}
\mathcal{D}_{fail}(m,d,r,s_k)
=
\sum_{f \in \mathcal{F}}
w_f
P
\left(
f
\middle|
\text{correct at } s_0
\rightarrow
\text{incorrect at } s_k
\right),
\end{equation}

where \(\mathcal{F}\) denotes the set of traversal failure categories and \(w_f\) denotes the severity weight associated with failure type \(f\).

\noindent \textit{Spatial Traversal Severity}, $\mathcal{D}_{spatial}(r,s_k)$, is directly represented as:

\begin{equation}
\mathcal{D}_{spatial}(r,s_k)
=
s_k,
\end{equation}

which measures the cumulative traversal corruption magnitude at traversal state \(s_k\).

\noindent \textit{Progressive Robustness Collapse}, $\mathcal{D}_{prog}(m,d,r,s_k)$, 
captures abrupt robustness collapse dynamics between consecutive traversal states, and is defined as:

\begin{equation}
\mathcal{D}_{prog}(m,d,r,s_k)
=
\max
\left(
0,
\frac{
A_{m,d,r,s_{k-1}}
-
A_{m,d,r,s_k}
}{
A_{m,d,r,s_{k-1}}+\epsilon
}
\right),
\end{equation}

where \(\epsilon > 0\) is a numerical stabilization constant.

\subsection{Robustness Corruption Index (RCI)}

For model $m$, dataset $d$, and perturbation condition $p$, RCI is defined as:

\begin{equation}
\begin{aligned}
\mathrm{RCI}(m,d,p) ={}&
\lambda_1 \mathcal{D}_{acc}(m,d,p)
+\lambda_2 \mathcal{D}_{cal}(m,d,p) \\
&+\lambda_3 \mathcal{D}_{ent}(m,d,p)
+\lambda_4 \mathcal{D}_{fail}(m,d,p).
\end{aligned}
\label{eq:rci}
\end{equation}

The RCI weights satisfy:
\vspace{-5pt}
\begin{equation}
\sum_{i=1}^{4}\lambda_i=1,
\qquad
\lambda_i \geq 0.
\end{equation}

\subsection{Traversal Robustness Corruption Index (T-RCI)}

For model \(m\), dataset \(d\), traversal direction \(r\), and traversal trajectory \(\tau_r\), we define T-RCI as:
\begin{equation}
\begin{aligned}
\mathrm{TRCI}(m,d,r)
={}&
\sum_{k=1}^{K}\gamma_k
\Big[
\alpha_1\mathcal{D}_{acc}(m,d,r,s_k)
\\
&+
\alpha_2\mathcal{D}_{ent}(m,d,r,s_k)
\\
&+
\alpha_3\mathcal{D}_{fail}(m,d,r,s_k)
\\
&+
\alpha_4\mathcal{D}_{spatial}(r,s_k)
\\
&+
\alpha_5\mathcal{D}_{prog}(m,d,r,s_k)
\Big].
\end{aligned}
\label{eq:trci}
\end{equation}

The aggregation weights satisfy:
\vspace{-5pt}
\begin{equation}
\sum_{i=1}^{5}\alpha_i = 1,
\qquad
\alpha_i \geq 0,
\end{equation}

and the traversal-step weighting coefficients satisfy:
\vspace{-5pt}
\begin{equation}
\sum_{k=1}^{K}\gamma_k = 1,
\qquad
\gamma_k \geq 0.
\end{equation}

To obtain direction-aggregated traversal robustness, T-RCI is further aggregated across traversal directions:
\begin{equation}
\mathrm{TRCI}(m,d)
=
\sum_{r \in \mathcal{R}}
\beta_r
\,
\mathrm{TRCI}(m,d,r),
\end{equation}

where the directional aggregation coefficients satisfy:
\vspace{-5pt}
\begin{equation}
\sum_{r \in \mathcal{R}} \beta_r = 1,
\qquad
\beta_r \geq 0.
\end{equation}

Unless otherwise stated, all aggregation weights are set uniformly ($\lambda_i=\tfrac{1}{4}$, $\alpha_i=\tfrac{1}{5}$, $\beta_r=\tfrac{1}{|\mathcal{R}|}$, and $\gamma_k=\tfrac{1}{K}$) to avoid introducing dataset-specific tuning and also ensure that no single robustness component may disproportionately influence the final score. Calibration is included in RCI because corruption-induced degradation is often accompanied by confidence miscalibration, whereas T-RCI instead incorporates the traversal progression instability term $D_{\mathrm{prog}}$, which captures abrupt robustness collapse across sequential traversal states. This design emphasizes the distinct failure dynamics induced by traversal perturbations, namely progressive information loss and spatial reasoning degradation. 

Uniform weighting was adopted as a neutral design choice because there is no established theoretical basis for assigning greater importance to one degradation mechanism over another, and introducing empirically tuned coefficients would have embedded additional assumptions into the robustness metrics. During metric development, we performed comprehensive ablation studies to evaluate multiple alternative weighting schemes, including accuracy-heavy, entropy-heavy, failure-heavy, equal-weight, and an ablation excluding the spatial severity component, to assess the sensitivity of the resulting robustness characterization. Across these configurations, the resulting T-RCI and RCI scores remained highly consistent with the submitted formulation (pairwise correlations of approximately $0.95$--$0.99$), which indicated that the qualitative robustness trends are largely insensitive to reasonable variations in the weighting strategy. Moreover, removing the spatial severity term $D_{\mathrm{spatial}}$ entirely still produced nearly identical robustness trends ($\rho \sim 0.95$), demonstrating that the reported conclusions are not driven by the model-independent traversal component but instead arise from the model-dependent degradation signals captured by the remaining terms. Given this observed stability, we retained the uniform weighting scheme as the simplest and least assumption-driven formulation.

\subsection{Different Types of Analyses Outputs}

The proposed framework produces a wide range of robustness analysis artifacts. These include: 
(1) overall RCI and T-RCI robustness scores; 
(2) directional fragility heatmaps for identifying spatial sensitivity patterns; 
(3) traversal robustness curves that characterize performance degradation across crop trajectories and severity levels; 
(4) failure taxonomy distributions that reveal dominant reasoning failure classes; and 
(5) comparative robustness rankings with Pareto-style analyses that summarize the trade-off between task performance and robustness across models and datasets.

\subsection{Failure Taxonomy Analysis}
To facilitate interpretable robustness analysis, corruption-induced prediction failures are categorized into four high-level reasoning failure domains: OCR, Spatial, Symbolic Reasoning, and Semantic.

\textit{OCR Failures} correspond to incorrect extraction or interpretation of textual information. 
\textit{Spatial Failures} arise from errors in geometric, structural, or diagrammatic reasoning. 
\textit{Symbolic Reasoning Failures} capture mistakes involving numerical, logical, or scientific reasoning processes. 
\textit{Semantic Failures} occur when models misinterpret contextual meaning or high-level relationships despite retaining partial visual understanding.

Each corruption-induced prediction error is assigned to its dominant failure category, enabling the analysis of how visual perturbations propagate into distinct reasoning breakdowns. Failure distributions are subsequently aggregated across perturbation conditions to identify dominant robustness vulnerabilities and characterize model-specific reasoning fragility. Failures that cannot be reliably assigned to one of the four primary domains are retained within a residual \textit{Other} category and reported separately for completeness. These can be attributed to failures arising from ambiguous, compound, hallucinated, unsupported, or otherwise non-attributable error mechanisms that cannot be confidently assigned to a single taxonomy category.

The proposed taxonomy is hierarchical rather than a direct flat classification. Individual failures are first analyzed using fine-grained corruption-specific cues (e.g., option-label loss, chemical-symbol parsing, diagram-text interpretation, molecule/bond interpretation, contextual misunderstanding, and reaction-mechanism reasoning), which are then consolidated into the four high-level reasoning domains (OCR, Spatial, Symbolic, and Semantic) and an ``Other'' category for ambiguous or uncategorized cases. The high-level taxonomy is designed to provide an interpretable summary of failure behavior rather than claim that all reasoning failures belong to a fixed set of fundamental categories. Moreover, failure categorization follows a predefined cue-based hierarchy rather than ad hoc assignment. Fine-grained reasoning cues are mapped to their corresponding high-level reasoning domains according to a fixed taxonomy, ensuring consistent categorization across all evaluated failures.
A detailed failure taxonomy is presented in Figure \ref{fig:finegrained}.

\section{Experimental Setup}

For evaluation, we focus on chemistry-oriented multimodal reasoning through \textit{MMJEE-Chemistry-MCQ} and \textit{ScienceQA-Chemistry}. The inclusion of \textit{MathVista-MCQ} allows us to investigate whether robustness degradation observed on chemistry-focused reasoning tasks generalizes to broader multimodal mathematical reasoning settings. For \textit{MMJEE-Chemistry-MCQ}, only English-language questions were retained during evaluation to ensure linguistic consistency between experiments and avoid additional variability introduced by prompt multilingual interpretation.

Multiple-choice question (MCQ) settings were used throughout the evaluation to enable stable and consistent robustness measurement across perturbation conditions. Compared to open-ended generation, MCQ evaluation reduces variability arising from unconstrained text generation and allows more reliable assessment of predictive degradation, entropy collapse, calibration drift, and corruption-induced reasoning transitions. The same answer parsing logic was used across all perturbation levels for a given dataset. For \textit{ScienceQA} and \textit{MathVista}, predictions were parsed as single uppercase option letters from the valid answer set. For \textit{MMJEE}, single-answer questions were parsed as one option letter, while multiple-answer questions were parsed as sorted unique option-letter sets. A prediction was marked correct only when the parsed answer exactly matched the ground-truth label.

Experiments were conducted using \underline{seven} modern vision-language models spanning both proprietary and open-source multimodal systems: \textit{GPT-4.1\cite{openai2025gpt41}, Claude Opus 4.7\cite{anthropic2025claudeopus47}, Gemini 3.5 Flash, Qwen2.5-VL-72B, InternVL2.5-8B, LLaVA-NeXT-Mistral-7B, and ChemVLM-8B}\cite{li2025chemvlm}.
These models were selected to provide architectural, scale, and training diversity across state-of-the-art multimodal systems. The evaluation includes both general-purpose vision-language models and a chemistry-specialized model, enabling robustness comparisons across different model families, reasoning capabilities, and domain specializations.

All models were evaluated using deterministic decoding to ensure consistent robustness measurements across perturbation conditions.
Temperature was fixed to 0 and stochastic sampling was disabled (\texttt{do\_sample=False}) whenever supported by the model API. Maximum generation length was fixed across experiments to avoid truncation-induced variability in answer extraction. For proprietary models (GPT-4.1, Claude Opus 4.7, and Gemini 3.5 Flash), images were provided directly through their respective multimodal APIs. Open-source models were evaluated using their official inference pipelines and processors without additional fine-tuning or adaptation. 
InternVL2.5-8B used the official InternVL image processor with images resized to the model's native input resolution (448$\times$448). LLaVA-NeXT-Mistral-7B was evaluated using the official LLaVA processor and image pre-processing pipeline. For API-based models, corrupted images were stored as PNG files and
submitted directly through the corresponding multimodal interfaces. GPT-4.1, Claude Opus 4.7, and Gemini 3.5 Flash therefore received
identical corrupted image inputs generated by the BRUCE perturbation pipeline.

Gaussian blur severity was controlled using kernel standard deviations ~\cite{hendrycks2019benchmarking}
$\sigma \in \{1,2,3,4\}$, while low-contrast perturbations employed
contrast factors $c \in \{0.30,0.50,0.70,0.90\}$.
Rotation perturbations used angles
$\theta \in \{5^\circ,10^\circ,15^\circ,20^\circ\}$, and center-crop
perturbations removed increasing fractions of image content using crop
ratios $p \in \{0.2,0.4,0.6,0.8\}$.

For a subset of evaluated models, reasoning traces generated during inference were additionally recorded alongside final predictions, which were used to investigate corruption-induced reasoning transitions, identify recurring failure mechanisms, and analyze how intermediate reasoning behavior changes as visual information progressively degrades.

\section{Results and Evaluation}

\begin{figure*}[!t]
\centering
\includegraphics[width=\textwidth]{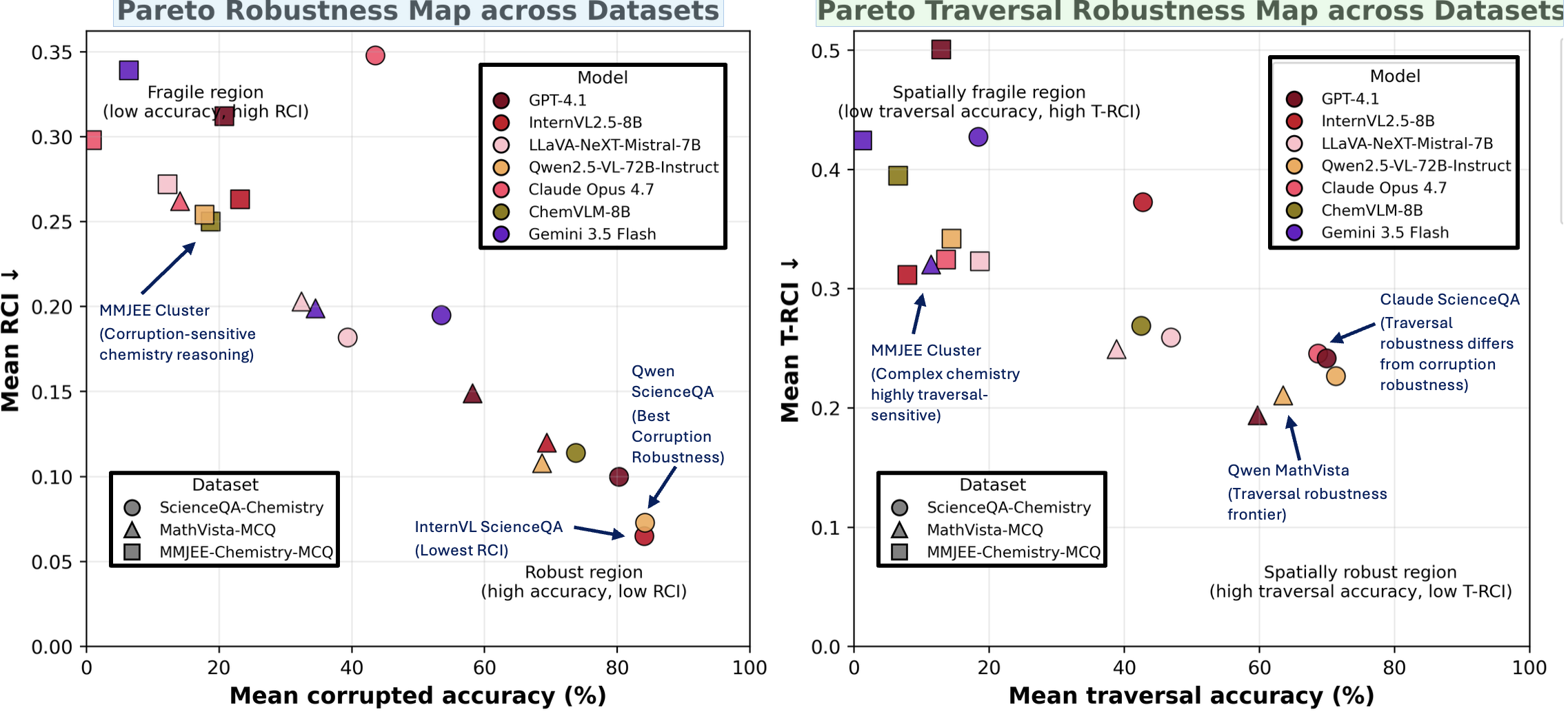}
\caption{
\textbf{Left:} \textit{Pareto Robustness Map} showing the trade-off between mean corrupted accuracy [\%] and Mean Robustness Corruption Index ($RCI_{\mu}$) across evaluated models and datasets. Lower-right regions indicate stronger robustness to corruption/perturbations.
\textbf{Right:} \textit{Pareto Traversal Robustness map} showing the trade-off between mean traversal accuracy [\%] and Mean Traversal Robustness Corruption Index ($T_{\mu}$). Lower-right regions indicate stronger robustness to progressive spatial information removal.
\textit{Colors denote models and marker shapes denote datasets.}
}
\label{fig:pareto_map}
\end{figure*}

\begin{table*}[t]
\centering
\small
\renewcommand{\arraystretch}{1.08}
\resizebox{\textwidth}{!}{
\begin{tabular}{lccccccccc}
\toprule
\textbf{Model} &
$\mathbf{A_{CL}}\uparrow$ &
$\mathbf{A_{CR}}\uparrow$ &
$\mathbf{RCI_{\mu}}\downarrow$ &
$\mathbf{RCI_{\max}}\downarrow$ &
$\mathbf{A_T}\uparrow$ &
$\mathbf{T_{\mu}}\downarrow$ &
$\mathbf{T_{\max}}\downarrow$ &
$\mathbf{W_D}$ &
$\mathbf{W_T}\downarrow$ \\
\midrule
\multicolumn{10}{l}{\textbf{MMJEE-Chemistry-MCQ}} \\
\midrule
ChemVLM-8B & 15.22 & 18.75 & \textbf{0.250} & 0.361 & 6.52 & 0.395 & 0.680 & RL & 0.479 \\
Claude Opus 4.7 & 0.00 & 0.82 & 0.298 & \textbf{0.300} & 13.59 & 0.324 & 0.561 & RL & 0.349 \\
GPT-4.1 & 21.74 & 20.79 & 0.312 & 0.541 & 12.91 & 0.501 & 0.707 & BU & 0.564 \\
Gemini 3.5 Flash & 6.52 & 6.39 & 0.339 & 0.520 & 1.22 & 0.424 & 0.770 & LR & 0.725 \\
InternVL2.5-8B & 21.74 & 23.19 & 0.263 & 0.311 & 7.86 & \textbf{0.311} & \textbf{0.366} & LR & \textbf{0.316} \\
LLaVA-NeXT-Mistral-7B & 10.87 & 12.23 & 0.272 & 0.414 & 18.61 & 0.323 & 0.670 & RL & 0.386 \\
Qwen2.5-VL-72B-Instruct & 13.04 & 17.80 & 0.254 & 0.384 & 14.40 & 0.342 & 0.563 & TD & 0.387 \\

\midrule
\multicolumn{10}{l}{\textbf{MathVista-MCQ}} \\
\midrule

GPT-4.1 & 60.19 & 58.16 & 0.149 & 0.184 & 59.69 & 0.194 & 0.304 & RL & \textbf{0.201} \\
Gemini 3.5 Flash & 32.00 & 34.50 & 0.199 & 0.254 & 11.38 & 0.320 & 0.470 & LR & 0.342 \\
InternVL2.5-8B & 71.85 & 69.35 & 0.120 & 0.195 & 34.44 & 0.538 & 0.632 & RL & 0.547 \\
LLaVA-NeXT-Mistral-7B & 32.41 & 32.41 & 0.203 & 0.203 & 38.84 & 0.249 & 0.310 & LR & 0.256 \\
Qwen2.5-VL-72B-Instruct & 69.63 & 68.62 & \textbf{0.108} & \textbf{0.173} & 63.50 & \textbf{0.211} & \textbf{0.305} & LR & 0.218 \\

\midrule
\multicolumn{10}{l}{\textbf{ScienceQA-Chemistry}} \\
\midrule
ChemVLM-8B & 77.66 & 73.74 & 0.114 & 0.220 & 42.49 & 0.269 & 0.504 & TD & 0.294 \\
Claude Opus 4.7 & 58.51 & 43.55 & 0.348 & 0.522 & 68.68 & 0.246 & 0.384 & LR & 0.282 \\
GPT-4.1 & 85.11 & 80.25 & 0.100 & 0.284 & 69.95 & 0.242 & 0.402 & RL & 0.306 \\
Gemini 3.5 Flash & 58.00 & 53.50 & 0.195 & 0.307 & 18.38 & 0.427 & 0.745 & RL & 0.526 \\
InternVL2.5-8B & 85.11 & 84.04 & \textbf{0.065} & 0.128 & 42.75 & 0.373 & 0.550 & RL & 0.473 \\
LLaVA-NeXT-Mistral-7B & 39.36 & 39.36 & 0.182 & 0.182 & 46.88 & 0.259 & \textbf{0.356} & RL & 0.276 \\
Qwen2.5-VL-72B-Instruct & \textbf{87.23} & \textbf{84.18} & 0.073 & \textbf{0.120} & \textbf{71.28} & \textbf{0.227} & 0.412 & RL & \textbf{0.269} \\

\bottomrule
\end{tabular}
}
\caption{
Benchmarking accuracy, corruption robustness, and traversal-corruption robustness of different VLMs across datasets.
}
\label{tab:unified_robustness_summary}
\vspace{2pt}

\begin{minipage}{\textwidth}
\footnotesize
$A_{CL}$ and $A_{CR}$ denote clean and mean corrupted accuracy, respectively.
$RCI_{\mu}$ and $RCI_{\max}$ denote mean and maximum RCI scores.
$A_T$ denotes mean traversal accuracy computed across non-zero traversal-crop severities.
$T_{\mu}$ and $T_{\max}$ denote mean and maximum T-RCI scores, respectively.
$W_D$ denotes the most fragile traversal direction, and $W_T$ denotes its corresponding T-RCI score.
Direction abbreviations: TD = Top-Down, BU = Bottom-Up, LR = Left-to-Right, and RL = Right-to-Left.
\textit{Lower RCI/T-RCI values and higher accuracy values indicate stronger robustness.}
\end{minipage}
\vspace{-10pt}
\end{table*}

\subsection{Benchmarking Accuracy, RCI and T-RCI of Different VLMs across Different Datasets}

Table \ref{tab:unified_robustness_summary} reports clean accuracy ($A_{CL}$), corrupted accuracy under perturbations ($A_{CR}$), and average and maximum values of the both robustness degradation metrics ($RCI_{\mu}$, $RCI_{max}$, $T_{\mu}$, and $T_{max}$) across all evaluated datasets and models. The last two columns also illustrate the most fragile traversal direction ($W_D$) and its corresponding T-RCI score ($W_T$) to provide more insights into the traversal robustness analysis.
In the following, we discuss the \textbf{key observations} from Table~\ref{tab:unified_robustness_summary} and Figure~\ref{fig:pareto_map} of Pareto-frontiers.

\textbf{O1:} 
\textit{Variations in the corrupted accuracy and robustness indices of different VLMs are highly dependent on the properties of the datasets.}
For instance, all models demonstrate comparatively stable clean and corrupted accuracies (e.g., above $80\%$ for some models) and low mean RCI (i.e., below $0.12$) for the ScienceQA-Chemistry dataset. 
In contrast, MMJEE-Chemistry-MCQ remains challenging even under
moderate perturbations, where corrupted accuracies frequently remain
below $25\%$, indicating substantially higher sensitivity of VLMs to visual
corruption. 
This dataset-dependent variation on VLMs' robustness and accuracy degradation under different types of perturbations is also reflected in the traversal Pareto analysis (Figure~\ref{fig:pareto_map}), where robustness for the ScienceQA-Chemistry dataset once again remains comparatively high, while substantially greater traversal sensitivity is visible for MMJEE-Chemistry-MCQ.

\textbf{O2:} 
\textit{Among the evaluated models, InternVL2.5-8B and Qwen2.5-VL-72B
consistently occupy the strongest robustness regime. 
}
On ScienceQA-Chemistry, InternVL2.5-8B achieves the lowest mean RCI
($0.065$) and high clean and corrupted accuracies 
($A_{CL}=85.11; A_{CR}=84.04$), while Qwen2.5-VL-72B obtains a similarly low mean RCI ($0.073$)
but provides the highest clean and corrupted accuracies ($A_{CL}=87.23; A_{CR}=84.18$).

This can be attributed to the higher number of parameters and stronger reasoning capabilities of Qwen2.5-VL-72B. Additionally, its training pipeline, incorporating extensive augmentation, OCR-rich data, and complex multimodal reasoning benchmarks,  may encourage greater robustness to localized information loss.
Across datasets, these models generally combine high retained accuracy with low degradation scores, suggesting more stable reasoning behavior under perturbations. 
Similarly, across traversal robustness analysis, Qwen2.5-VL-72B also occupies the traversal Pareto frontier and achieves the lowest mean T-RCI on both MathVista-MCQ and ScienceQA-Chemistry, while InternVL2.5-8B attains the lowest mean T-RCI on MMJEE-Chemistry-MCQ. InternVL's robustness is particularly notable because it is achieved despite substantially fewer parameters than Qwen2.5-VL-72B, suggesting that effective vision-language alignment and visual grounding may play a larger role than model scale alone~\cite{qwen25vl2025}. 

\textbf{O3:} GPT-4.1 and Gemini 3.5 Flash exhibit two very distinct failure modes on the MMJEE-Chemistry-MCQ. GPT-4.1 achieves one of the strongest clean accuracies on it ($A_{CL}=21.74$), indicating comparatively stronger chemistry reasoning capability under clean conditions. However, this advantage does not translate into traversal robustness since the model exhibits the highest traversal degradation among all evaluated models ($T_{\mu}=0.501$, $T_{\max}=0.707$) and only moderate traversal accuracy ($A_T=12.91$), demonstrating that performance deteriorates substantially as chemically relevant regions are progressively removed. This suggests that GPT-4.1 relies heavily on preserving complete visual evidence and chemically informative structures, achieving correctness through effective visual grounding rather than robustness to localized information loss. In contrast, Gemini 3.5 Flash exhibits both low clean accuracy ($A_{CL}=6.52\%$) and the lowest traversal accuracy ($A_T=1.22\%$), together with severe traversal degradation ($T_{\mu}=0.424$, $T_{\max}=0.770$). These results illustrate that GPT-4.1 is capable but spatially fragile, whereas Gemini is limited in both chemistry competence and traversal robustness, which is consistent with the prior studies~\cite{chemiq2025,cui2025chem_olympiad} on Gemini-family models. 

\textbf{O4:} 
\textit{Clean-task performance does not always translate into corruption robustness.} 
For example, Claude Opus 4.7 exhibits competitive performance on several clean benchmarks yet records the highest mean RCI on ScienceQA-Chemistry ($0.348$), substantially exceeding the degradation observed for InternVL2.5-8B and Qwen2.5-VL-72B. Moreover, Claude Opus 4.7 exhibits a unique zero clean accuracy in MMJEE-Chemistry-MCQ, suggesting that benchmark difficulty rather than corruption-induced degradation is the dominant failure factor in this setting. Unexpectedly, Claude Opus 4.7 exhibits a very different behavior under traversal perturbations, since it remains relatively competitive on ScienceQA-Chemistry despite poor robustness. \textit{This suggests that corruption robustness and traversal robustness capture distinct failure mechanisms.}

Claude Opus 4.7's 0\% clean accuracy on MMJEE-Chemistry-MCQ is unexpected at first glance. However, interpreting this observation as evidence of either an implementation error or a general deficiency in chemistry reasoning is not correct. First, the result was reproduced across repeated runs under the same prompting and parsing pipeline, suggesting that it is not attributable to mere sampling noise. Second, prior work has documented that frontier models, including Claude, can exhibit benchmark-specific failures arising from interactions between alignment mechanisms, symbolic reasoning requirements, and evaluation structure. For example, the Robust Reasoning Benchmark reports systematic failures of Claude variants on certain symbol-manipulation tasks \cite{golikov2026rrb}, while Anthropic has documented sensitivity to benchmark-like prompt formats \cite{anthropic2026evalawareness}. Third, Claude Opus 4.7 remains comparatively competitive on ScienceQA-Chemistry and exhibits substantially different traversal behavior, indicating that its performance degradation is not uniform across chemistry benchmarks. Taken together, these observations suggest that the MMJEE result reflects a benchmark-dependent interaction rather than a simple monotonic relationship between overall model capability and benchmark accuracy. We therefore refrain from attributing the anomaly to a single cause and leave a more detailed investigation of model-specific effects to future work.

\textbf{O5:}
\textit{Interestingly, the chemistry-specialized ChemVLM-8B does not exhibit a clear robustness advantage on the chemistry-oriented benchmarks}. While ChemVLM-8B was evaluated exclusively on MMJEE-Chemistry-MCQ and ScienceQA-Chemistry, its robustness metrics remain comparable to, and in some cases weaker than, several general-purpose VLMs. One possible explanation is that chemistry specialization primarily improves chemical reasoning capability rather than robustness to degraded visual inputs. Since corruption and traversal perturbations fundamentally alter the visual evidence available to the model, resilience may depend more heavily on visual grounding, multimodal alignment, and exposure to diverse visual conditions during pretraining than on chemistry-specific supervision alone. Moreover, another contributing factor may be model scale;  ChemVLM-8B possesses substantially fewer parameters than models such as Qwen2.5-VL-72B, after all. Consequently, the gains obtained from domain specialization may be insufficient to offset the broader multimodal representations and robustness acquired through large-scale pretraining.

\textbf{O6:} 
\textit{Figure \ref{fig:pareto_map} reveals that models with similar retained accuracies can exhibit substantially different robustness scores which can be attributed to differences in their model sizes, data augmentation pipelines, and training loops.} Several model--dataset pairs occupy comparable regions along the accuracy axis while remaining separated along the RCI or T-RCI dimensions. This observation indicates that predictive accuracy alone does not fully characterize robustness. As a result, robustness-aware metrics such as RCI and T-RCI provide complementary information by quantifying degradation dynamics which remain hidden under conventional accuracy-based evaluation.

\textbf{O7:} \textit{Furthermore, traversal perturbations can serve as a coarse diagnostic of spatial evidence utilization by revealing which image regions contribute most strongly to the final prediction.} This provides a complementary perspective on model reasoning by identifying which parts of an image are most important for successful prediction. If removing a small region causes a substantial performance drop, it suggests that the model's reasoning may depend on a limited subset of visual evidence rather than a holistic understanding of the scene and context, a phenomenon previously associated with shortcut feature reliance and incomplete multimodal grounding \cite{xing2025visualquality,ross2017right}.

\textbf{O8:} \textit{VLM's robustness evaluations for the MathVista-MCQ dataset lie between ScienceQA-Chemistry and MMJEE-Chemistry-MCQ.} 
While degradation levels are observed to be lower than those observed on MMJEE-Chemistry-MCQ, they are generally higher than those witnessed on ScienceQA-Chemistry. 
This suggests that mathematical and geometric visual reasoning is moderately resilient to corruption and traversal perturbations, but still depends on localized diagrammatic evidence that can get disrupted under progressive spatial removal.

\textbf{O9:} \textit{Table~\ref{tab:unified_robustness_summary} further demonstrates that robustness degradation depends not only on perturbation severity but also on traversal direction.}
Several models exhibit markedly different T-RCI scores across traversal trajectories, indicating that information within scientific visual reasoning tasks is not distributed uniformly across the image plane. 
Moreover, the recurring appearance of Right-to-Left as the most fragile direction on ScienceQA-Chemistry suggests the presence of shared directional vulnerabilities that persist across architectures. 
These findings motivate future investigation into spatial information localization and directional reasoning biases in VLMs.
\vspace{-5pt}

\begin{figure*}[!t]
\centering
\includegraphics[width=\textwidth]
{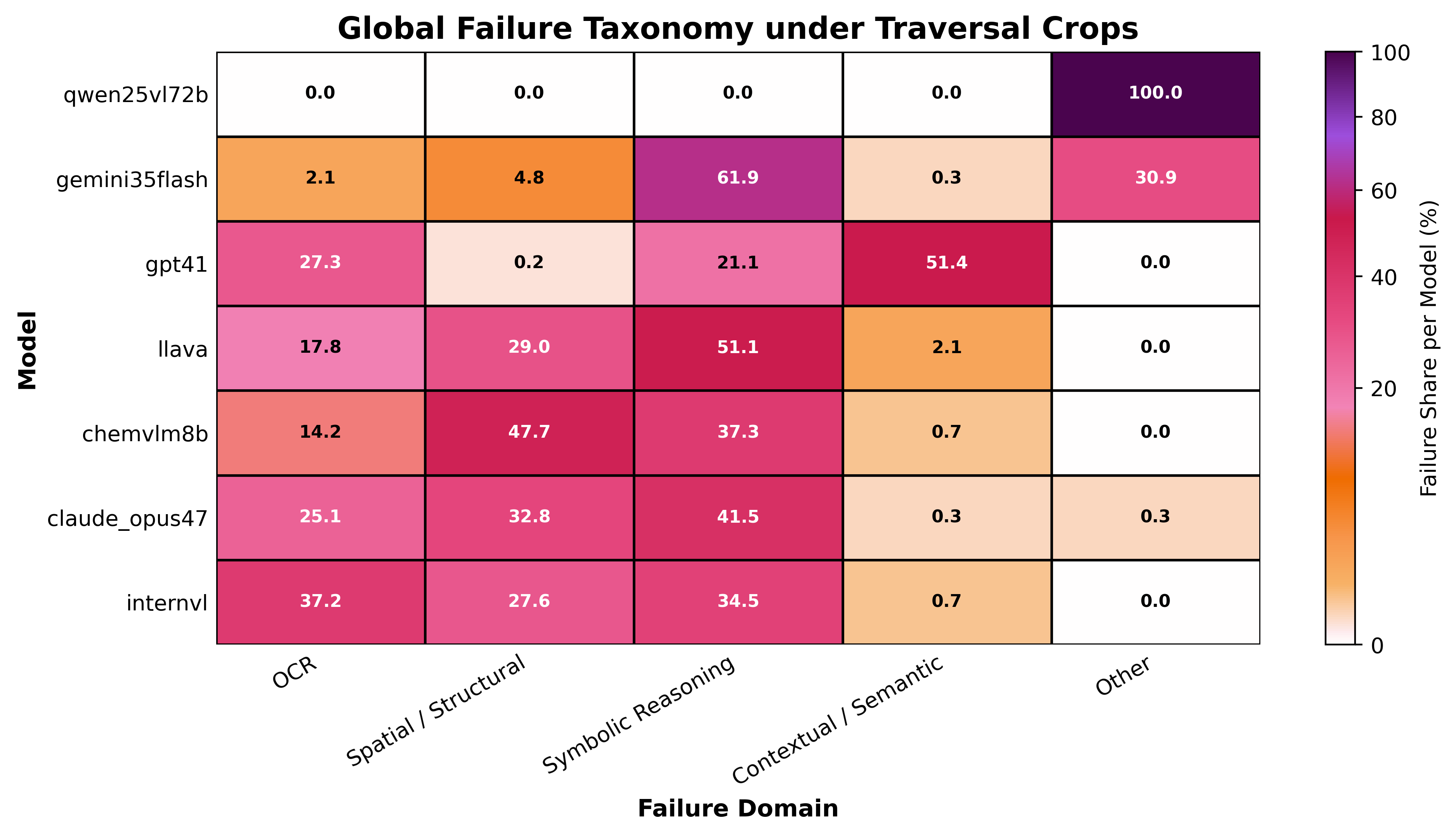}
\caption{
Global traversal-induced failure taxonomy across evaluated vision-language models under directional spatial corruption.
}
\label{fig:global_failure_taxonomy}
\end{figure*}

\subsection{Understanding the Impact of Traversal-induced Degradation}

To better understand the underlying causes of traversal-induced degradation, we categorize model failures into four interpretable reasoning domains: \textit{OCR}, \textit{Spatial}, \textit{Symbolic Reasoning}, and \textit{Semantic}, with \textit{Other} being the outlier. Figure~\ref{fig:global_failure_taxonomy} summarizes the distribution of these failure types aggregated across traversal directions and severity levels. From this figure, we derive the following key reasoning observations.

\textbf{R1:}
\textit{A clear trend emerges across nearly all evaluated models: symbolic reasoning failures dominate the error distribution.} 
For Gemini 3.5 Flash, LLaVA-NeXT-Mistral-7B, ChemVLM-8B, Claude Opus 4.7, and InternVL2.5-8B, symbolic failures constitute the largest fraction of traversal-induced errors, as progressive spatial information loss primarily disrupts multi-step reasoning processes rather than low-level visual perception. 

In many cases, models are still capable of recognizing visual content but fail to correctly manipulate numerical, logical, or scientific relationships required to reach the final answer.

\textbf{R2:}
\textit{The prevalence of symbolic failures provides a potential explanation for the elevated T-RCI values observed under directional traversals.} 
As visual information is progressively removed from specific image regions, models often preserve partial perceptual understanding while losing the relational context required for reasoning~\cite{zeiler2014visualizing,fong2017interpretable}. Consequently, performance degradation is frequently driven by failures in reasoning composition rather than complete visual recognition collapse.

\textbf{R3:}
\textit{While symbolic reasoning dominates for most models, GPT-4.1 exhibits a markedly different failure profile.} 
More than half of its traversal-induced failures are categorized as contextual/semantic, indicating that GPT-4.1 retains comparatively stronger symbolic reasoning under traversal degradation but becomes increasingly vulnerable to disruptions affecting global context and scene-level understanding. One possible explanation is that GPT-4.1 retains comparatively strong symbolic reasoning capabilities even under substantial traversal perturbations. Rather than exhibiting direct reasoning collapse, the model often continues to generate coherent reasoning chains from the remaining visual evidence. However, when traversal removes semantically important regions, GPT-4.1 may form an incomplete or incorrect interpretation of the scene before reasoning begins. Consequently, many failures manifest as contextual or semantic misunderstandings rather than symbolic reasoning errors. Contrastingly, several smaller models more frequently experience breakdowns in the reasoning process itself, leading to a larger proportion of symbolic failures instead. This divergence suggests that different VLM architectures may rely on distinct reasoning pathways when confronted with partial visual information.

\textbf{R4:}
The observed failure distributions further align with the benchmark-level trends reported in the robustness and traversal analyses. In particular, \textit{chemistry-oriented reasoning tasks appear especially susceptible to symbolic degradation}. Many chemistry questions require the interpretation of structured visual elements, symbolic notation, numerical relationships, and multi-step scientific reasoning chains. As traversal perturbations progressively remove image content, these dependencies become increasingly fragile, leading to elevated degradation observed on MMJEE-Chemistry-MCQ and chemistry-focused benchmarks.

\textit{Overall, the taxonomy analysis demonstrates that traversal robustness is not solely a perception problem.} Instead, directional information loss frequently propagates into higher-level reasoning failures, with symbolic reasoning emerging as the dominant failure mode across most evaluated VLMs.

\subsection{Spatial Traversal Robustness Analysis}

\begin{figure*}[!t]
\centering
\includegraphics[width=\textwidth]{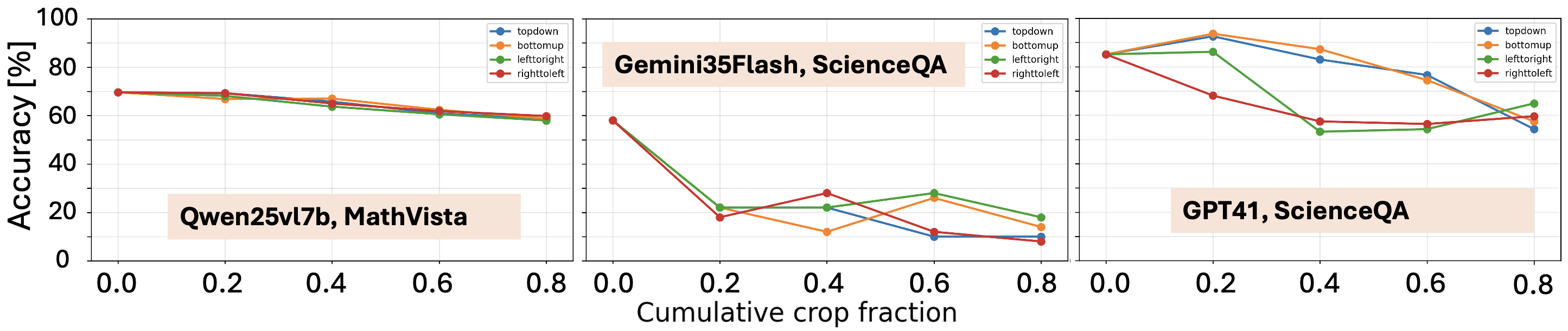}
\caption{
Spatial traversal robustness curves under progressive directional cropping. Accuracy is measured as increasing portions of the image are removed from four traversal directions (top-down, bottom-up, left-to-right, and right-to-left). Qwen2.5-VL-7B shows stable degradation; GPT-4.1 exhibits progressive degradation; and Gemini 3.5 Flash shows severe traversal sensitivity.
}
\label{fig:Spatial Traversal}
\end{figure*}

Figure~\ref{fig:Spatial Traversal} shows spatial traversal robustness curves under progressive directional cropping, from which we derive the following key observations.

\textbf{S1:} 
Based on the shape of the curve, the individual traversal behaviors of the evaluated VLMs can be broadly grouped into three robustness regimes: (1) stable degradation, (2) progressive degradation, and (3) severe traversal sensitivity. Variations in these robustness regimes can be attributed to differences in model architectures, size, data augmentation, and training pipelines, and their resulting  internal contextual reasoning capabilities.

\textbf{S2:} 
\textit{Qwen2.5-VL-7B on MathVista exhibits a stable degradation profile, where performance gradually decreased as larger portions of the image were removed.} 
Despite substantial information loss, the model maintains relatively high accuracy throughout the traversal trajectory, suggesting that its reasoning process is distributed across multiple visual regions rather than being dependent on a single localized cue. 

\textbf{S3:} 
\textit{The degradation is not strictly monotonic, but rather, several models exhibit temporary accuracy improvements at intermediate crop severities before subsequently declining.} What this behavior suggests is that certain image regions may introduce distracting or misleading visual information and their removal can occasionally simplify the reasoning process. Similar non-monotonic robustness effects have been observed in previous studies of corruption robustness and distribution shift, where 
perturbations can sometimes improve model performance even though the perception image fidelity is reduced ~\cite{xing2025visualquality,
hendrycks2021many,sui2025benchc}.

\textbf{S4:} 
\textit{Substantial differences are witnessed along traversal directions.} 
GPT-4.1 exhibits noticeably different degradation trajectories between top-down and right-to-left traversals, which indicates that robustness depends not only on the amount of information removed but also on the spatial location from which information is removed, consistent with prior observations that different image regions contribute unequally to model predictions and reasoning behavior~\cite{zeiler2014visualizing,fong2017interpretable}. \textit{This observation motivates the use of proposed T-RCI metric for robustness benchmarking, which explicitly captures directional reasoning fragility that would otherwise be obscured by direction-agnostic corruption metrics.}
\vspace{5pt}

\section{Detailed Discussion on Further Observations}

\begin{figure*}[!t]
\centering
\includegraphics[width=\textwidth]{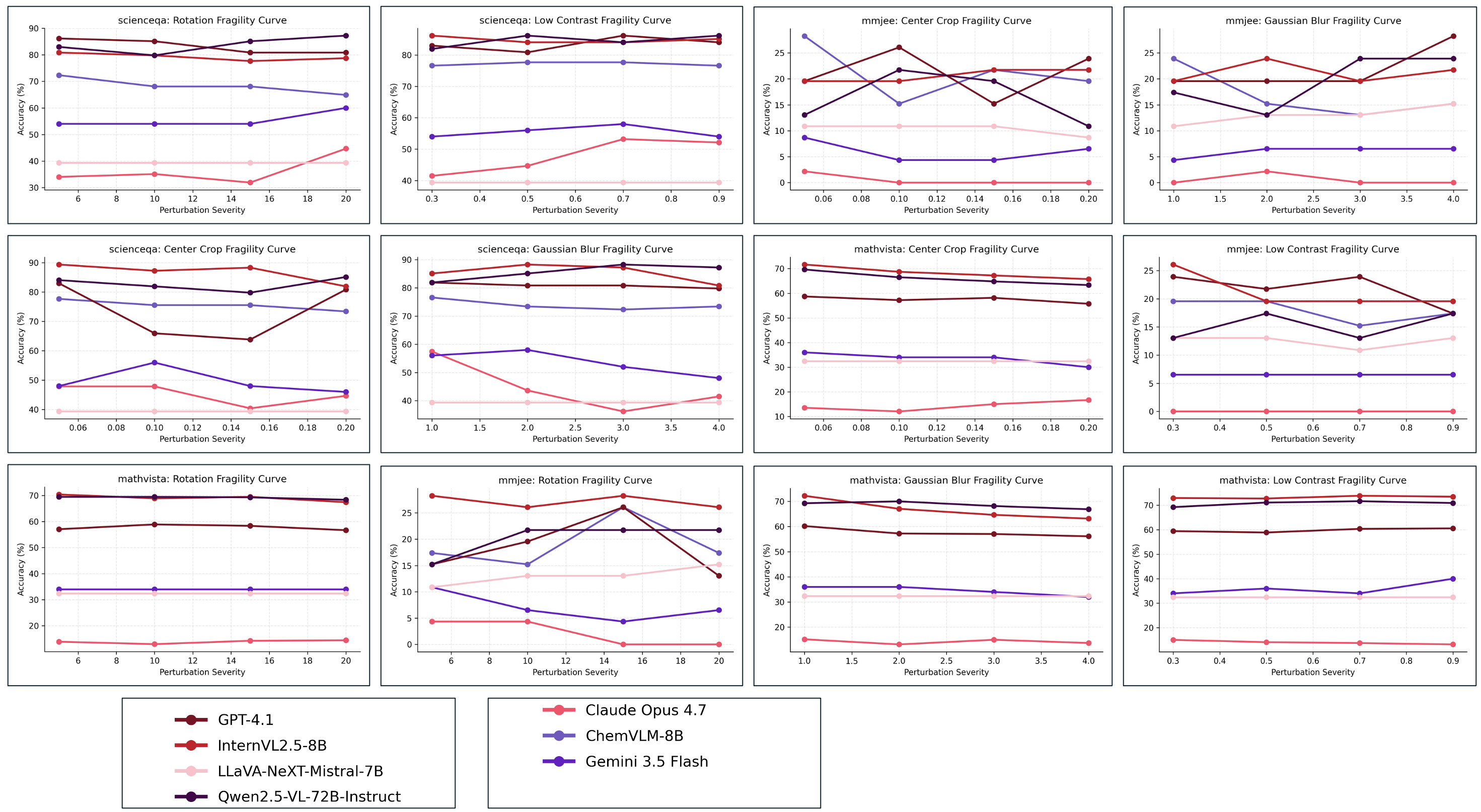}
\caption{
Fragility curves plotted for all models across datasets.
}
\label{fig:frag}
\end{figure*}

MathVista exhibits substantially different robustness characteristics compared to the chemistry-oriented benchmarks. Across nearly all perturbation categories, model rankings remain relatively stable and accuracy degradation is generally less severe than that observed on MMJEE-Chemistry-MCQ. This suggests that many MathVista questions remain partially solvable even when moderate visual information is degraded, whereas chemistry reasoning often depends on preserving fine-grained symbolic and structural details. 

Figure~\ref{fig:frag} shows one notable observation that is the consistently strong performance of InternVL2.5-8B and Qwen2.5-VL-72B across nearly all MathVista perturbation settings. GPT-4.1 also demonstrates strong robustness on MathVista, particularly under center-crop perturbations, with only modest degradation as information is progressively removed.

The MathVista Gaussian blur curves further support this interpretation. While blur substantially reduces fine-grained visual details, several models maintain relatively stable performance. This suggests that many MathVista reasoning tasks depend more heavily on coarse structural relationships than on pixel-level symbolic fidelity. In contrast, chemistry questions involving molecular structures, reaction diagrams, or symbolic notation often become significantly more vulnerable once fine structural information is degraded.

The MathVista fragility curves further illustrate that traversal robustness and conventional corruption robustness capture distinct failure mechanisms. Several models remain comparatively stable under blur and low-contrast perturbations yet exhibit noticeably larger degradation under center-crop perturbations. This indicates that preserving global image quality alone is insufficient when task-critical spatial evidence is selectively removed, further motivating traversal-based robustness evaluation and T-RCI.

\section{Traversal-Induced Failure Case Study}
\begin{figure*}[!t]
\centering
\includegraphics[width=\textwidth]{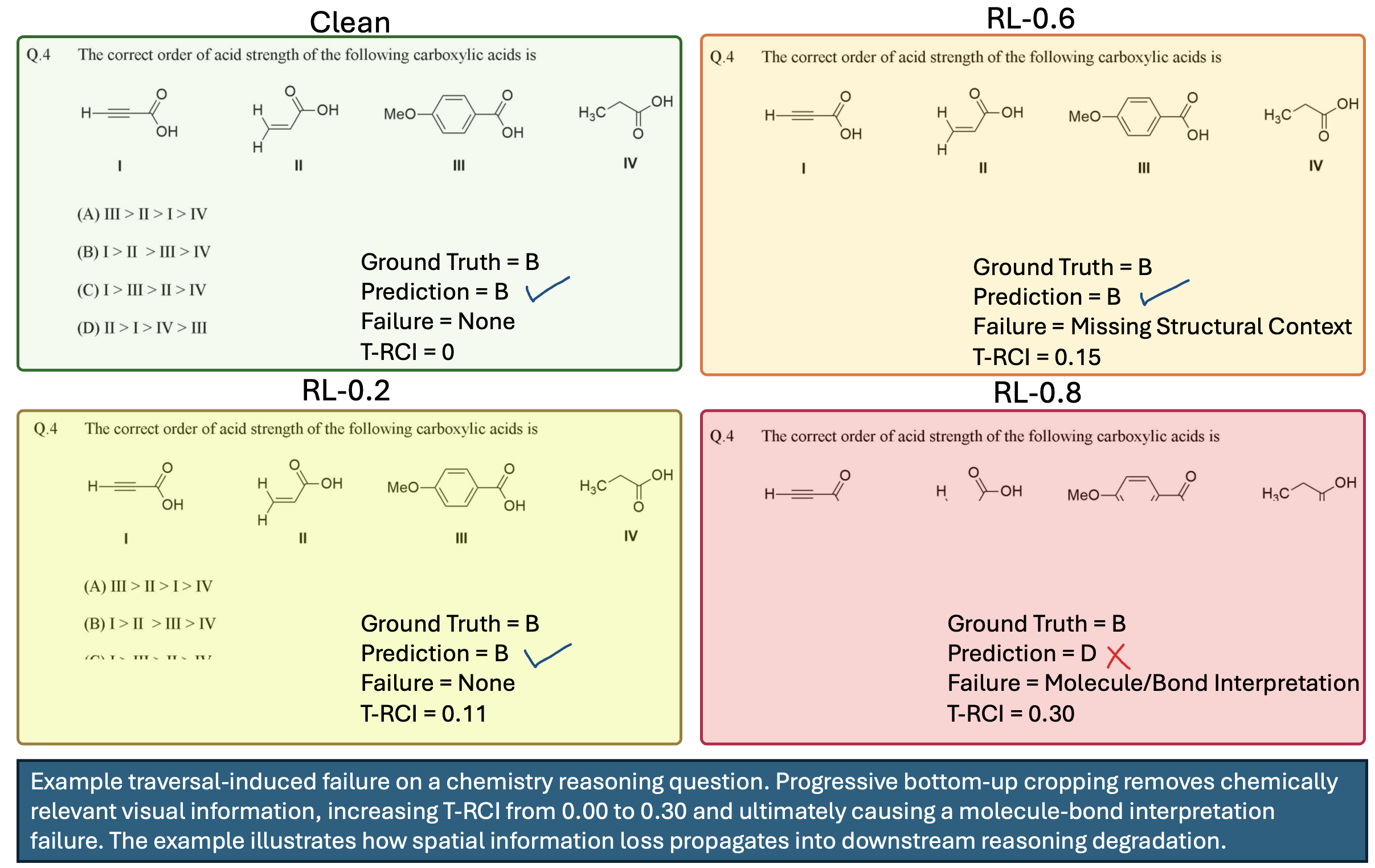}
\caption{
Example traversal-induced failure on a chemistry reasoning question.
}
\label{fig:4ww}
\end{figure*}

Figure \ref{fig:4ww} presents a representative example of traversal-induced reasoning degradation on a chemistry question from MMJEE-Chemistry-MCQ. Observe that the clean image contains four molecular structures whose relative acid strengths must be compared. Under the clean condition, the model correctly predicts option B. As progressively larger portions of the image are removed through bottom-up traversal cropping, the available visual evidence becomes increasingly incomplete. At mild to mid traversal severities (e.g., \textbf{RL-0.2} and \textbf{RL-0.6}), the model continues to produce the correct answer despite partial information loss. However, the T-RCI score increases from 0.00 to 0.15, indicating growing instability even before an observable prediction error occurs. This highlights a very important property of T-RCI: \textit{robustness degradation can emerge prior to accuracy degradation}, allowing latent failure signals to be detected before a complete reasoning collapse occurs.

At \textbf{RL-0.8}, substantial portions of the molecular structures have now been removed. In particular, chemically relevant bond configurations, functional groups, and structural cues become partially unavailable. Consequently, the model changes its prediction from the correct answer (B) to an incorrect answer (D). The corresponding failure is categorized as a \emph{molecule/bond interpretation failure}, since the traversal operation prevents accurate reconstruction of the structural relationships required for acidity comparison.

Under the clean image, the model was able to identify relevant molecular structures, compare electron-withdrawing and electron-donating effects, and estimate the relative stability of the conjugate bases. These chemical relationships collectively support the correct ranking corresponding to option B. As traversal severity increases, portions of the molecular structures become unavailable, which reduces the model's ability to accurately identify substituent effects and molecular connectivity. Rather than explicitly recognizing that information is missing, the model continues reasoning using incomplete visual evidence. The resulting reasoning chain remains internally coherent but is grounded on an incorrect structural interpretation, ultimately producing the incorrect prediction. This example highlights a common traversal-induced failure mode in which visual grounding errors propagate into downstream symbolic reasoning errors.

\section{Fine-Grained Traversal-Induced Failure Analysis}

\begin{figure*}[!t]
\centering
\includegraphics[width=\textwidth]{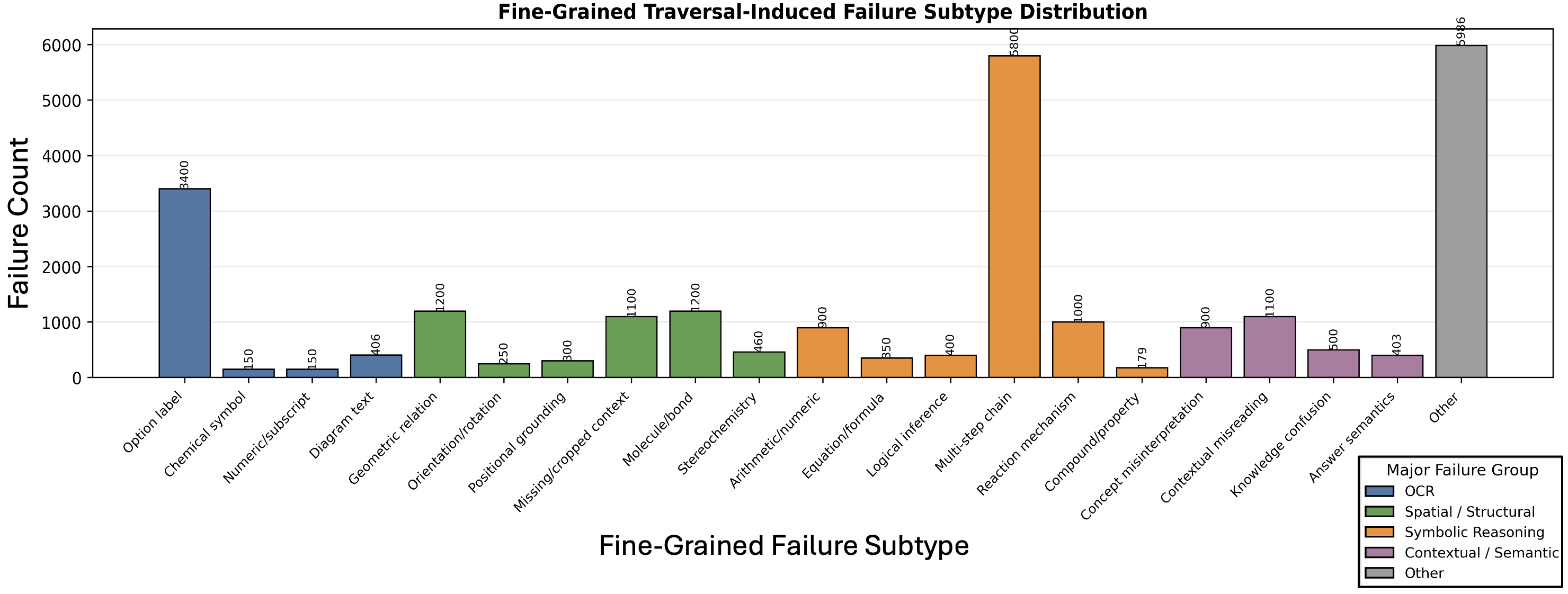}
\caption{
Global traversal-induced failure taxonomy across evaluated vision-language models under directional spatial corruption.
}
\label{fig:finegrained}
\end{figure*}

Figure~\ref{fig:finegrained} presents the distribution of fine-grained failure subtypes aggregated across all models, datasets, traversal directions, and traversal severities. The taxonomy provides a more detailed view of how spatial information loss propagates into downstream reasoning failures.
\subsection{OCR Failures}
\textit{There are 3400 instances of Option Label:}  
This category represents failures where traversal removes or partially obscures answer labels, option identifiers, or localized textual markers required for answer selection. Such failures occur frequently because many benchmark questions encode critical information through option labels rather than the visual content itself. When traversal crops remove these regions, the model may correctly reason about the underlying content but fail to map the reasoning outcome to the correct answer choice. 
 
\textit{Moving on, 150 instances of Chemical Symbol Parsing exist:} 
These failures arise when traversal removes individual atomic symbols, functional group annotations, or reaction labels. Although relatively infrequent, errors of this kind can fundamentally alter chemical interpretation because a single missing symbol may completely change molecular identity.

\textit{Numeric/Subscript Parsing counts for 150 instances:}  
These failures occur when numerical coefficients, stoichiometric subscripts, exponents, or quantitative annotations are partially removed. Even minor corruption of numerical information can invalidate otherwise correct reasoning chains.

\textit{Next, Diagram Text appears on 406 instances:}  
Traversal frequently removes labels embedded inside scientific diagrams, charts, molecular structures, and reaction schemes. Since many scientific benchmarks rely on integrated visual-textual reasoning, missing diagram text often creates ambiguity in downstream interpretation.

\subsection{Spatial and Structural Failures}

\textit{Geometric Relation appears 1200 times:} 
These failures occur when traversal disrupts relative spatial relationships between visual entities. Examples include geometric diagrams, graph structures, positional comparisons, and object arrangements. Since traversal explicitly alters spatial visibility, geometric failures naturally emerge as a major failure mode.

\textit{Orientation/Rotation makes up 250 instances:} 
Certain questions require interpreting directional orientation, molecular configuration, stereochemical arrangement, or spatial alignment. Partial visibility often prevents accurate recovery of orientation-dependent information.

\textit{Positional Grounding makes up 300 instances:} 
These failures arise when models lose the ability to accurately localize relevant regions after traversal removes contextual anchors. The model may identify the correct concept but associate it with the wrong spatial location.

\textit{Missing/Cropped Context (accumulates at 1100 instances:} 
This category captures failures directly induced by traversal itself. The model loses access to visually relevant regions required for complete interpretation, resulting in incomplete evidence and degraded reasoning performance.

\textit{Molecule/Bond Interpretation is observed at 1200 instances:} 
This represents one of the dominant chemistry-specific failure modes. Traversal frequently removes chemically meaningful structures such as bond types, substituent groups, ring structures, or reaction components. Because chemical reasoning often depends on subtle structural cues, even small information loss can substantially alter molecular interpretation.

\textit{Stereochemistry (460 instances):} 
Such failures arise because traversal removes stereochemical markers such as wedges, dashes, chiral centers, or geometric configuration indicators. Such information is often spatially localized and therefore highly vulnerable to directional cropping.

\subsection{Symbolic Reasoning Failures}

\textit{Arithmetic/Numeric Reasoning (900 instances):} 
These failures occur when numerical reasoning chains become unstable due to incomplete visual evidence. Missing values, truncated tables, or partially visible numerical relationships frequently lead to incorrect calculations.

\textit{Equation/Formula Reasoning failures constitute 350 instances:} 
Scientific equations and symbolic expressions often span multiple regions of the image. Traversal may remove variables, operators, or critical formula components, causing downstream reasoning errors.

\textit{Logical Inference makes up 400 instances:} 
These failures arise when models can no longer correctly infer relationships between multiple pieces of visual evidence after traversal disrupts information flow.

\textit{Multi-Step Chain Failure (5800 instances):} 
These constitute the most prevalent explicit reasoning failure category. Scientific reasoning tasks frequently require sequential inference chains where each step depends on previous observations. Traversal-induced information loss often causes early-stage errors that propagate through the entire reasoning process, leading to complete reasoning collapse. The dominance of this category suggests that robustness degradation frequently manifests through cascading inference failures rather than isolated prediction errors.

\textit{Reaction Mechanism failures are observed at 1000 instances):} 
Chemical reaction pathways often require interpreting multiple interacting structures simultaneously. Partial removal of reactants, intermediates, catalysts, or products can disrupt mechanistic reasoning.

\textit{Compound Property Reasoning (179 instances):}  
Failures involve incorrect prediction of molecular properties, acidity trends, polarity, reactivity, stability, or related chemistry concepts due to incomplete structural information.

\subsection{Contextual and Semantic Failures}

\textit{Concept Misinterpretation is common at 900 instances:} 
The model observes the remaining visual evidence but constructs an incorrect semantic understanding of the underlying concept.

\textit{Contextual Misreading occurs in 1100 instances:} 
These failures occur when traversal removes supporting context required for accurate interpretation. The model often generates internally coherent reasoning chains but reasons from incomplete evidence.

\textit{Knowledge Confusion occurred at 500 instances:} 
The model retrieves or applies incorrect background knowledge when traversal removes disambiguating visual cues.

\textit{Answer Semantics (403 instances):} 
These failures occur when the model understands the underlying concept but incorrectly maps its reasoning to the final answer semantics or response formulation.

\subsection{Other Failures}

\textit{Other (5986 instances):} 
The \textit{Other} category contains residual failures that do not cleanly map onto a single fine-grained subtype. These can be attributed to failures arising from ambiguous, compound, hallucinated, unsupported, or otherwise non-attributable error mechanisms that cannot be confidently assigned to a single taxonomy category. The large size of this category highlights the complexity of traversal-induced degradation and suggests that robustness failures often arise through interactions between multiple reasoning components rather than isolated failure mechanisms.

\subsection{Overall Observations}
Several noteworthy trends emerge. First, symbolic reasoning failures, particularly multi-step chain failures, dominate the taxonomy, indicating that traversal-induced degradation frequently propagates through reasoning pipelines. Second, chemistry-specific failures such as molecule/bond interpretation and reaction mechanism reasoning appear prominently, confirming that scientific reasoning tasks are highly sensitive to localized structural information loss. Third, contextual misreading and concept misinterpretation remain substantial contributors, suggesting that many failures arise from incomplete scene understanding rather than pure reasoning collapse. Finally, the persistence of a large \textit{Other} category indicates that traversal perturbations often trigger complex failure cascades involving simultaneous degradation of visual grounding, semantic interpretation, and symbolic reasoning.

\section{Discussion and Conclusion}

In this study, we present BRUCE, a comprehensive framework for evaluating the robustness of VLMs under both visual corruption perturbations and progressive spatial traversal perturbations. We introduce the \textit{Robustness Corruption Index (RCI)} and \textit{Traversal Robustness Corruption Index (T-RCI)}, two complementary metrics to jointly quantify performance degradation and  deterioration of underlying reasoning behavior. 
Our extensive benchmarking campaigns and analysis of the results reveal the following key observations, which provide important guidelines for designing domain-specific VLMs for scientific problems:

\vspace{5pt}
\begin{itemize}
    \item Models that maintain strong performance under visual corruption perturbations do not necessarily preserve stable reasoning behavior. Across multiple settings, accuracy alone was insufficient to characterize robustness, thereby requiring RCI and T-RCI.

\vspace{5pt}    
    \item Progressive removal of directional information exposes vulnerabilities that remain hidden otherwise. Spatial traversal induces systematic failures, which demonstrated that robustness to corruption and robustness to information loss represent distinct aspects.

    \vspace{5pt}
    \item Many models exhibit noticeably different traversal degradation regimes, ranging from stable degradation to severe traversal sensitivity, highlighting substantial variation in how visual evidence is utilized during reasoning.

    \vspace{5pt}
    \item The reasoning capabilities of VLMs for the chemistry problems frequently depend on localized structural, symbolic, and spatial cues whose removal disrupts the reasoning process disproportionately.

    \vspace{5pt}
    \item Fine-grained failure analysis using traversal perturbations illustrate that robustness failures often arise from reasoning collapse rather than purely perceptual degradation.

\end{itemize}

\bibliographystyle{IEEEtran}
\bibliography{egbib}

@article{liu2023mmbench,
  title={MMBench: Is Your Multi-modal Model an All-around Player?},
  author={Liu, Yuan and Duan, Haodong and Zhang, Wenqi and others},
  journal={arXiv preprint arXiv:2307.06281},
  year={2023}
}

@inproceedings{hendrycks2021many,
  title={The Many Faces of Robustness: A Critical Analysis of Out-of-Distribution Generalization},
  author={Hendrycks, Dan and Basart, Steven and Mu, Norman and others},
  booktitle={Proceedings of the IEEE/CVF International Conference on Computer Vision (ICCV)},
  pages={8340--8349},
  year={2021}
}

@inproceedings{ross2017right,
  title={Right for the Right Reasons: Training Differentiable Models by Constraining their Explanations},
  author={Ross, Andrew Slavin and Hughes, Michael C. and Doshi-Velez, Finale},
  booktitle={Proceedings of the Twenty-Sixth International Joint Conference on Artificial Intelligence (IJCAI)},
  pages={2662--2670},
  year={2017},
  doi={10.24963/ijcai.2017/371}
}

@article{xing2025visualquality, title={Demystifying the Visual Quality Paradox in Multimodal Large Language Models}, author={Xing, Shuo and Guo, Lanqing and Hua, Hongyuan and others}, journal={CoRR}, volume={abs/2506.15645}, year={2025} }

@article{yue2023mmmu,
  title={MMMU: A Massive Multi-discipline Multimodal Understanding and Reasoning Benchmark for Expert AGI},
  author={Yue, Xiang and Ni, Yuansheng and Zhang, Kai and Zheng, Tianyu and Liu, Ruoqi and Zhang, Ge and Stevens, Samuel and Jiang, Dongfu and Ren, Weiming and Sun, Yuxuan and Wei, Cong and Yu, Botao and Yuan, Ruibin and Sun, Renliang and Yin, Ming and Zheng, Boyuan and Yang, Zhenzhu and Liu, Yibo and Huang, Wenhao and Sun, Huan and Su, Yu and Chen, Wenhu},
  journal={arXiv preprint arXiv:2311.16502},
  year={2023}
}

@article{chen2024mmstar,
  title={MMStar: Are We on the Right Way for Evaluating Large Vision-Language Models?},
  author={Chen, Lin and Zhao, Yilun and Liu, Yuxuan and others},
  journal={arXiv preprint arXiv:2403.20330},
  year={2024}
}

@article{arXiv:2412.19794,
  title={MVTamperBench: Evaluating Robustness of Vision-Language Models},
  author={Agarwal, Amit and Panda, Srikant and Charles, Angeline and others},
  journal={arXiv preprint arXiv:2412.19794},
  year={2024}
}

@article{becker2026causally,
  title={A Causally Grounded Taxonomy for Image Degradation Robustness Evaluation},
  author={Becker, Stefan and Weiss, Simon and H{\"u}bner, Wolfgang and Arens, Michael},
  journal={arXiv preprint arXiv:2605.15906},
  year={2026},
  url={https://arxiv.org/abs/2605.15906}
}

@inproceedings{lu2022learn,
  title={Learn to Explain: Multimodal Reasoning via Thought Chains for Science Question Answering},
  author={Lu, Pan and Mishra, Swaroop and Xia, Tony and Qiu, Liang and Chang, Kai-Wei and Zhu, Song-Chun and Tafjord, Oyvind and Clark, Peter and Kalyan, Ashwin},
  booktitle={Advances in Neural Information Processing Systems (NeurIPS)},
  year={2022},
  url={https://arxiv.org/abs/2209.09513}
}

@inproceedings{mukherjee2025mmjee,
  title={mmJEE-Eval: A Bilingual Multimodal Benchmark for Evaluating Scientific Reasoning in Vision-Language Models},
  author={Mukherjee, Arka and Ghosh, Shreya},
  booktitle={Findings of the Association for Computational Linguistics: IJCNLP 2025},
  year={2025},
  url={https://arxiv.org/abs/2511.09339}
}

@article{lu2023mathvista,
  title={MathVista: Evaluating Mathematical Reasoning of Foundation Models in Visual Contexts},
  author={Lu, Pan and Bansal, Hritik and Xia, Tony and Liu, Jiacheng and Li, Chunyuan and Hajishirzi, Hannaneh and Cheng, Hao and Chang, Kai-Wei and Galley, Michel and Gao, Jianfeng},
  journal={arXiv preprint arXiv:2310.02255},
  year={2023},
  url={https://arxiv.org/abs/2310.02255}
}

@article{liu2024rbench,
  title={R-Bench: Are Your Large Multimodal Models Robust to Real-world Corruptions?},
  author={Li, Chunyi and Zhang, Jianbo and Zhang, Zicheng and Wu, Haoning and Tian, Yuan and Sun, Wei and Lu, Guo and Liu, Xiaohong and Min, Xiongkuo and Lin, Weisi and Zhai, Guangtao},
  journal={arXiv preprint arXiv:2410.05474},
  year={2024}
}

@misc{anthropic2025claudeopus47,
  title={Introducing Claude Opus 4.7},
  author={{Anthropic}},
  year={2025},
  howpublished={\url{https://www.anthropic.com/news/claude-opus-4-7}},
  note={Accessed 2026}
}

@inproceedings{agarwal2025vqa,
  title={Visual Robustness Benchmark for Visual Question Answering (VQA)},
  author={Ishmam, Md Farhan and Tashdeed, Ishmam and Saadat, Talukder Asir and Ashmafee, Md Hamjajul and Kamal, Abu Raihan Mostofa and Hossain, Md. Azam},
  booktitle={Proceedings of the IEEE/CVF Winter Conference on Applications of Computer Vision (WACV)},
  year={2025}
}

@article{liu2023llava,
  title={Visual Instruction Tuning},
  author={Liu, Haotian and Li, Chunyuan and Wu, Qingyang and Lee, Yong Jae},
  journal={arXiv preprint arXiv:2304.08485},
  year={2023}
}

@article{dai2023instructblip,
  title={InstructBLIP: Towards General-purpose Vision-Language Models with Instruction Tuning},
  author={Dai, Wenliang and Li, Junnan and Li, Dongxu and Tiong, Anthony and Zhao, Junqi and Wang, Weisheng and Sebe, Nicu and Hoi, Steven},
  journal={arXiv preprint arXiv:2305.06500},
  year={2023}
}

@article{bai2023qwenvl,
  title={Qwen-VL: A Frontier Large Vision-Language Model with Versatile Abilities},
  author={Bai, Jinze and Bai, Shuai and Yang, Shusheng and Wang, Sheng and Tan, Silu and Wang, Ke and Huang, Xinghao and others},
  journal={arXiv preprint arXiv:2308.12966},
  year={2023}
}

@article{wang2023cogvlm,
  title={CogVLM: Visual Expert for Pretrained Language Models},
  author={Wang, Yizhong and Mishra, Swaroop and Alabi, Jesujoba and others},
  journal={arXiv preprint arXiv:2311.03079},
  year={2023}
}

@article{chen2024internvl,
  title={InternVL 2.0: Scaling up Vision Foundation Models and Aligning for Generic Visual-Linguistic Tasks},
  author={Chen, Zhe and Wang, Weiyun and Cao, Yue and Liu, Zheng and others},
  journal={arXiv preprint arXiv:2403.20377},
  year={2024}
}

@article{li2025chemvlm,
  title={ChemVLM: Exploring the Power of Multimodal Large Language Models in Chemistry Area},
  author={Li, Junxian and Zhang, Di and Wang, Xunzhi and others},
  journal={Proceedings of the AAAI Conference on Artificial Intelligence},
  volume={39},
  number={1},
  pages={415--423},
  year={2025},
  doi={10.1609/aaai.v39i1.32020}
}

@misc{openai2025gpt41,
  title        = {Introducing GPT-4.1 in the API},
  author       = {{OpenAI}},
  year         = {2025},
  howpublished = {\url{https://openai.com/index/gpt-4-1/}},
  note         = {Accessed 2026}
}

@inproceedings{guo2024molpuzzle,
  title={Can LLMs Solve Molecule Puzzles? A Multimodal Benchmark for Molecular Structure Elucidation},
  author={Guo, Kehan and Zhang, Yifei and Liu, Shuo and others},
  booktitle={Advances in Neural Information Processing Systems (NeurIPS) Datasets and Benchmarks Track},
  year={2024}
}

@article{zhou2025chemtable,
  title={Benchmarking Multimodal LLMs on Recognition and Understanding over Chemical Tables},
  author={Zhou, Yitong and Cheng, Mingyue and Mao, Qingyang and others},
  journal={arXiv preprint arXiv:2506.11375},
  year={2025},
  url={https://arxiv.org/abs/2506.11375}
}

@inproceedings{ying2024mmtbench,
  title={MMT-Bench: A Comprehensive Multimodal Benchmark for Evaluating Large Vision-Language Models Towards Multitask AGI},
  author={Ying, Kaining and Meng, Fanqing and Wang, Jin and others},
  booktitle={Proceedings of the 41st International Conference on Machine Learning (ICML)},
  pages={57116--57198},
  year={2024}
}

@article{usama2025corruptions,
  title={Analysing the Robustness of Vision-Language-Models to Common Corruptions},
  author={Usama, Muhammad and Asim, Syeda Aishah and Ali, Syed Bilal and Wasim, Syed Talal and Mansoor, Umair Bin},
  journal={CoRR},
  volume={abs/2504.13690},
  year={2025}
}

@article{sui2025benchc,
  title={Benchmarking Corruption Robustness of LVLMs: A Discriminative Benchmark and Robustness Alignment Metric},
  author={Sui, Xiangjie and Li, Songyang and Zhu, Hanwei and Chen, Baoliang and Fang, Yuming and Sun, Xin},
  journal={arXiv preprint arXiv:2511.19032},
  year={2025}
}

@article{recht2018cifar,
  title={Do CIFAR-10 Classifiers Generalize to CIFAR-10?},
  author={Recht, Benjamin and Roelofs, Rebecca and Schmidt, Ludwig and Shankar, Vaishaal},
  journal={arXiv preprint arXiv:1806.00451},
  year={2018}
}

@inproceedings{recht2019imagenet,
  title={Do ImageNet Classifiers Generalize to ImageNet?},
  author={Recht, Benjamin and Roelofs, Rebecca and Schmidt, Ludwig and Shankar, Vaishaal},
  booktitle={Proceedings of the 36th International Conference on Machine Learning (ICML)},
  pages={5389--5400},
  year={2019}
}

@inproceedings{zeiler2014visualizing,
  title={Visualizing and Understanding Convolutional Networks},
  author={Zeiler, Matthew D. and Fergus, Rob},
  booktitle={Computer Vision -- ECCV 2014},
  pages={818--833},
  year={2014},
  publisher={Springer}
}

@article{chemiq2025,
  title={Assessing the Chemical Intelligence of Large Language Models},
  author={Runcie, Nicholas T. and others},
  journal={arXiv preprint arXiv:2505.07735},
  year={2025}
}

@article{qwen25vl2025,
  title={Qwen2.5-VL Technical Report},
  author={Qwen Team and others},
  journal={arXiv preprint arXiv:2502.13923},
  year={2025}
}

@article{cui2025chem_olympiad,
  title={Evaluating Large Language Models on Multimodal Chemistry Olympiad Exams},
  author={Cui, Yiming and Yao, Xin and Qin, Yuxuan and Li, Xin and Wang, Shijin and Hu, Guoping},
  journal={Communications Chemistry},
  volume={8},
  number={1},
  pages={402},
  year={2025},
  publisher={Nature}
}

@inproceedings{fong2017interpretable,
  title={Interpretable Explanations of Black Boxes by Meaningful Perturbation},
  author={Fong, Ruth and Vedaldi, Andrea},
  booktitle={Proceedings of the IEEE International Conference on Computer Vision (ICCV)},
  pages={3429--3437},
  year={2017}
}

@inproceedings{hendrycks2019benchmarking,
  title={Benchmarking Neural Network Robustness to Common Corruptions and Perturbations},
  author={Hendrycks, Dan and Dietterich, Thomas},
  booktitle={ICLR},
  year={2019}
}

@article{golikov2026rrb,
  author  = {Pavel Golikov and Evgenii Opryshko and Gennady Pekhimenko and Mark C. Jeffrey},
  title   = {Robust Reasoning Benchmark},
  journal = {arXiv preprint arXiv:2604.08571},
  year    = {2026},
  doi     = {10.48550/arXiv.2604.08571},
  url     = {https://arxiv.org/abs/2604.08571}
}

@misc{anthropic2026evalawareness,
  author       = {{Anthropic}},
  title        = {Eval Awareness in Claude Opus 4.6's BrowseComp Performance},
  howpublished = {\url{https://www.anthropic.com/engineering/eval-awareness-browsecomp}},
  year         = {2026},
  month        = mar,
  note         = {Anthropic Engineering Blog}
}

\newpage
\newpage
\pagebreak 
\clearpage

\begin{IEEEbiography}[{\includegraphics[width=1in,height=1.25in,clip,keepaspectratio]{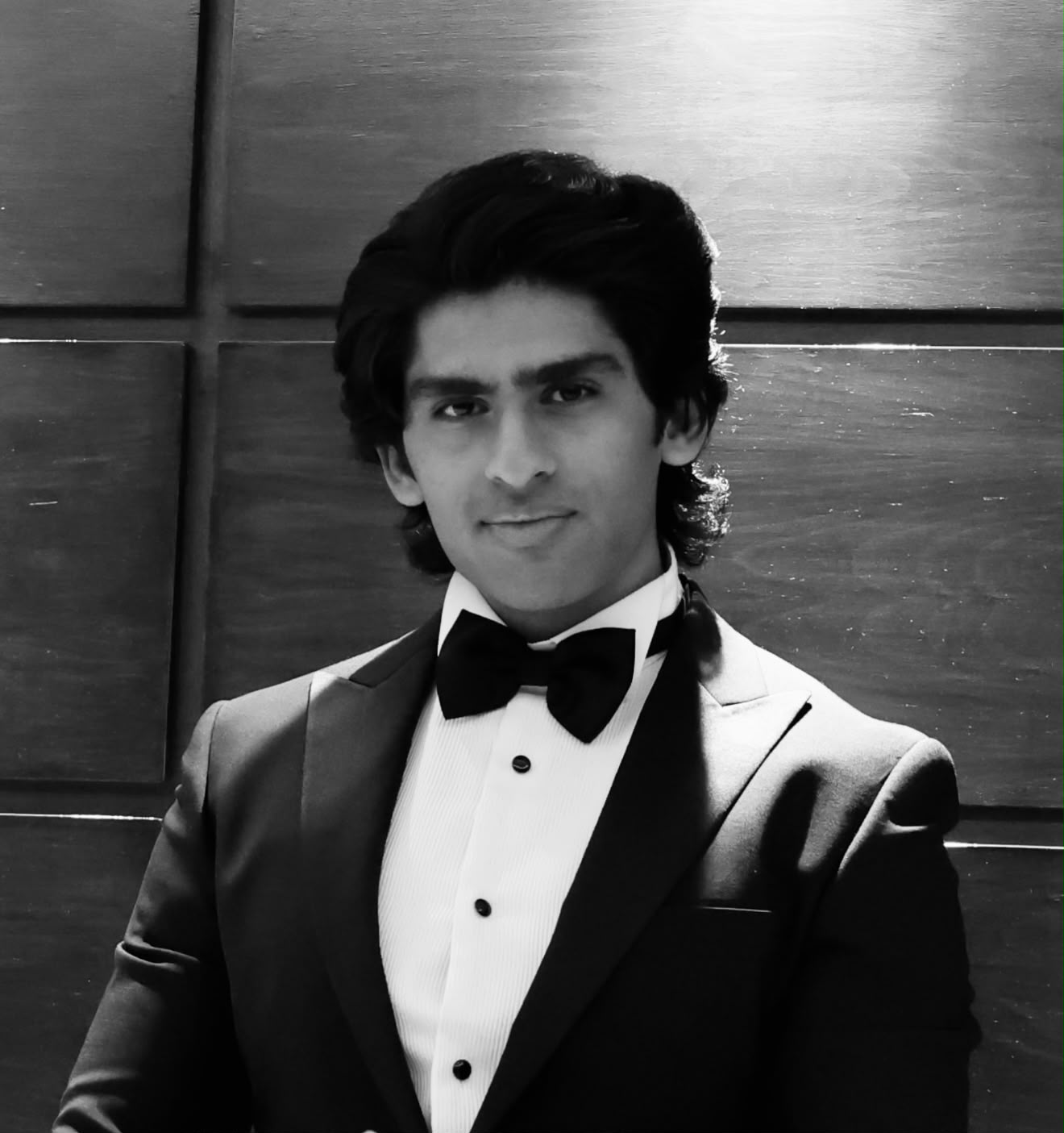}}]{Saim Rehman}
(Student Member, IEEE) is currently pursuing his undergraduate degree in Computer Engineering at New York University Abu Dhabi (NYUAD). His research interests are in AI \& machine learning hardware, electronic design automation, system-level design, brain-inspired computing, EdgeAI, tinyML, quantum machine learning, and multi-UAV systems. He was the recipient of the Best Researcher Award 2024, awarded by eBRAIN Lab, NYUAD. 
\end{IEEEbiography}

\vspace{-30pt}

\begin{IEEEbiography}[{\includegraphics[width=1in,height=1.25in,clip,keepaspectratio]{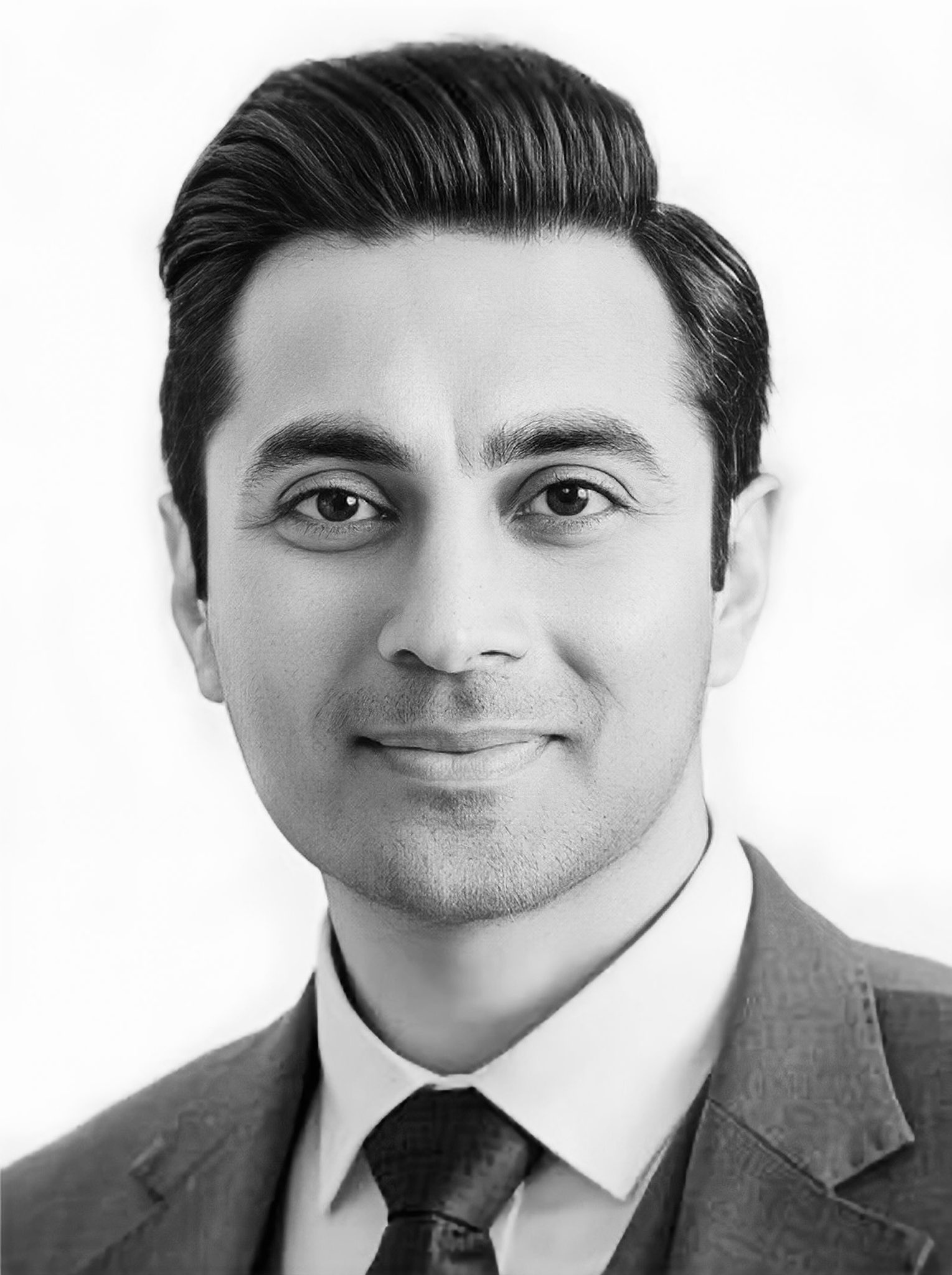}}]{Muhammad Shafique} (M’11 - SM’16) received the Ph.D. degree in computer science from the Karlsruhe Institute of Technology (KIT), Germany, in 2011. In Oct.2016, he joined the Faculty of Informatics at TU Wien, Vienna, Austria as a Full Professor of Computer Architecture and Robust, Energy-Efficient Technologies. Since Sep.2020, Dr. Shafique is with the New York University (NYU), where he is currently a Full Professor and the director of eBRAIN Lab and iCAS Lab at the NYU-Abu Dhabi in UAE, and a Global Network Professor at the Tandon School of Engineering, NYU-New York City in USA. He is also a Co-PI/Investigator in multiple NYUAD Centers on Cybersecurity, Quantum Computing, AI \& Robotics, and Smart Cities. His research interests are in AI \& machine learning hardware and system-level design, brain-inspired computing, EdgeAI, tinyML, machine learning security and privacy, quantum machine learning, cognitive autonomous systems, wearable healthcare, AI for healthcare/medical imaging, energy-efficient systems, robust computing, hardware security, emerging technologies, electronic design automation, FPGAs, MPSoCs, and embedded systems. The researched technologies and tools are deployed in application use cases from IoT, Smart CPS, Healthcare and Robotics domains.
Dr. Shafique has given several Keynotes, Invited Talks, and Tutorials, as well as organized many special sessions at premier venues. He has served as the PC Chair, General Chair, Track Chair, and PC member for several prestigious IEEE/ACM conferences. Dr. Shafique holds one U.S. patent, and has (co-)authored 10 Books, 25+ Book Chapters, 450+ papers in premier journals and conferences, and 200+ archive articles. He received the 2015 ACM/SIGDA Outstanding New Faculty Award, the AI-2000 Chip Technology Most Influential Scholar Awards (2020, 2022, 2023; Honorable Mention 2024, 2025), the ASPIRE AARE Research Excellence Award in 2021, six gold medals, several best paper awards and nominations at prestigious conferences, several HiPEAC paper awards, and multiple competition awards. He is a senior member of the IEEE and IEEE Signal Processing Society (SPS), and a senior member of the ACM, SIGARCH, SIGDA, SIGBED, and HIPEAC.

\end{IEEEbiography}

\EOD

\end{document}